\documentclass[sn-mathphys-num]{sn-jnl}

\usepackage{graphicx}%
\usepackage{multirow}%
\usepackage{amsmath,amssymb,amsfonts}%
\usepackage{amsthm}%
\usepackage{mathrsfs}%
\usepackage[title]{appendix}%
\usepackage{xcolor}%
\usepackage{textcomp}%
\usepackage{manyfoot}%
\usepackage{booktabs}%
\usepackage{algorithm}%
\usepackage{algorithmicx}%
\usepackage{algpseudocode}%
\usepackage{listings}%

\usepackage{bbding}
\usepackage{enumitem}
\usepackage{dsfont}
\usepackage{array}
\usepackage{amsmath}
\usepackage{amssymb}
\usepackage{tabularx}
\usepackage{hyperref}

\theoremstyle{thmstyleone}%
\theoremstyle{thmstyletwo}%

\theoremstyle{thmstylethree}%

\begin{document}

\title[Large Language Model Enhanced Knowledge Representation Learning: A Survey]{Large Language Model Enhanced Knowledge Representation Learning: A Survey}

\author*[1]{\fnm{Xin} \sur{Wang}} \email{wangx@tju.edu.cn}
\author[1]{\fnm{Zirui} \sur{Chen}}
\author*[2]{\fnm{Haofen} \sur{Wang}}\email{carter.whfcarter@gmail.com}
\author[3]{\fnm{Leong Hou} \sur{U}}
\author[1]{\fnm{Zhao} \sur{Li}}
\author[1]{\fnm{Wenbin} \sur{Guo}}

\affil[1]{\orgdiv{College of Intelligence and Computing}, \orgname{Tianjin University}, \orgaddress{\city{Tianjin}, \country{China}}}
\affil[2]{\orgdiv{College of Design and Innovation}, \orgname{Tongji University}, \orgaddress{\city{Shanghai}, \country{China}}}
\affil[3]{\orgdiv{Faculty of Science and Technology}, \orgname{University of Macao}, \orgaddress{\city{Macao}, \country{China}}}

\abstract{Knowledge Representation Learning (KRL) is crucial for enabling applications of symbolic knowledge from Knowledge Graphs (KGs) to downstream tasks by projecting knowledge facts into vector spaces. Despite their effectiveness in modeling KG structural information, KRL methods are suffering from the sparseness of KGs. The rise of Large Language Models (LLMs) built on the Transformer architecture presents promising opportunities for enhancing KRL by incorporating textual information to address information sparsity in KGs. LLM-enhanced KRL methods, including three key approaches, encoder-based methods that leverage detailed contextual information, encoder-decoder-based methods that utilize a unified Seq2Seq model for comprehensive encoding and decoding, and decoder-based methods that utilize extensive knowledge from large corpora, have significantly advanced the effectiveness and generalization of KRL in addressing a wide range of downstream tasks. This work provides a broad overview of downstream tasks while simultaneously identifying emerging research directions in these evolving domains.}


\maketitle

\section{Introduction}
Triples in Knowledge Graphs (KGs) are an effective form of structured knowledge representation, consisting of an entity, a relation, and an object, where the object can be either another entity or a literal. This structured fact is crucial for various downstream tasks such as link prediction \cite{link_prediction} (e.g., predicting "Albert Einstein" is associated with "Physics" through the relation "field of study"), triple classification \cite{triple_classification} (e.g., verifying if "Paris is the capital of France" is true), and relation classification \cite{relation_classification} (e.g., classifying "wrote" as the relation connecting "J.K. Rowling" and "Harry Potter"). To better leverage the symbolic knowledge contained in KGs for these downstream tasks, various Knowledge Representation Learning (KRL) approaches have been developed. Notable methods like TransE \cite{transe}, RESCAL \cite{rescal}, and R-GCN \cite{r-gcn} focus on embedding information (entities, relations, etc.) into low-dimensional vector spaces.

Despite being effective in modelling KG structural information, these methods suffer from the sparseness of KGs. Specifically, some KRL models are trained to preserve the inherent structure of KGs, favoring entities that are highly connected. According to the research, it is widely observed that the degrees of entities in KGs approximately follow the power-law distribution, resulting in a long tail populated with massive unpopular entities of low degrees \cite{long_tail}. Consequently, due to information sparsity, the performance of KRL methods tends to degrade when processing long-tail entities. In short, the sparseness of KGs significantly impacts the effectiveness of of representing low-degree entities, posing an existing challenge.

To tackle the above challenges, a promising solution is to enhance KRL approaches with large language models. Large Language Models (LLMs), built on the Transformer architecture \cite{transformer}, have gained significant popularity in the NLP field due to their remarkable performance. By being pre-trained on vast text corpora, LLMs demonstrate profound content comprehension and rich real-world knowledge, which can be utilized to address the information sparsity in KGs by incorporating textual information such as entity descriptions, offering potential opportunities for better knowledge representations.

Nowadays, the potential of LLM-enhanced KRL methods has attracted increasing attention across both academia and industry. In the stages of LLM development, these models exhibit an increase in parameters and real-world knowledge, progress from good to strong understanding capabilities, and transition from task-specific to more generalized outputs. This evolution can be illustrated through the development of three Transformer architectures. Accordingly, LLM-enhanced KRL methods can be categorized into three types based on these architectures. In encoder-based methods, the textual context is fully leveraged to capture detailed contextual information for knowledge representation. In encoder-decoder-based methods, a seq2seq model is utilized to intuitively and simply meet all requirements through efficient encoding and decoding. In decoder-based methods, extensive knowledge from large corpora is fully harnessed for downstream tasks.

Few existing surveys have explored the enhancement of KRL using LLM. With respect to KRL, Ge et al. \cite{ge_survey} provide a comprehensive overview of the two main branches in KRL: distance-based and semantic matching-based approaches, with few references to LLM enhancement. Biswas et al. \cite{biswas_survey} give an overview of the advancements in KRL by incorporating features such as semantics, multi-modal, temporal, and multilingual aspects, while rarely addressing LLM empowerment. Cao et al. \cite{cao_survey} classify and analyze KRL models from the perspectives of mathematical spaces, specifically through algebraic, geometric, and analytical perspective, excluding considerations related to LLM augmentation. During this period, Pan et al. \cite{pan_shirui_survey} mainly focus on the integration of LLMs and KGs from the perspective of general frameworks, KG-enhanced LLMs, LLM-augmented KGs, and Synergized LLMs + KGs. They demonstrate how to leverage their respective advantages in a complementary manner, with a minor portion dedicated to LLM-enhanced KRL.

To the best of our knowledge, this paper is the first to present a detailed categorization of LLM-enhanced KRL methods, conduct comprehensive reviews of downstream tasks, and identify emerging directions in these rapidly evolving fields. The contributions of this survey are summarized as follows:

\begin{itemize}[leftmargin=25px]
    \item \textbf{Comprehensive Survey of Techniques.} This survey provides a comprehensive overview of LLM-enhanced KRL techniques, examining the integration of encoder-based, encoder-decoder-based, and decoder-based methods with LLMs.
    \item \textbf{Review of Existing Evaluations.} 
    We systematically compile and analyze experimental results from recent studies, providing a consolidated perspective on their respective merits and challenges across various downstream tasks.
    \item \textbf{Future Directions.} The survey explores the foundational principles of LLM-enhanced KRL and proposes six promising avenues for future research.
\end{itemize}

The rest of this survey is organized as follows. Section \ref{Sec_2} outlines the pre-LLM KRL approaches, including translation model and semantic matching model. These methods laid the groundwork for modern approaches and provide essential context for understanding subsequent advancements. Section \ref{Sec_3} provides a detailed illustration of LLM-enhanced KRL methods, categorized by three types of Transformer: encoder-based, encoder-decoder-based, and decoder-based methods. Each category is discussed along with its representative works and the developments in subsequent research. Section \ref{Sec_4} introduces widely used datasets and the discussion of various evaluation approaches, offering insights into the relative strengths and weaknesses of different methods. Section \ref{Sec_5} introduces six potential future directions, from evolving knowledge representation to addressing downstream tasks with KRL+LLM. Finally, Section \ref{Sec_6} summarizes this paper.

\section{Pre-LLM KRL} \label{Sec_2}
Prior to the emergence of LLMs, KRL approaches primarily focused on modeling the structural information of KGs, which are known as the pre-LLM KRL methods. Depending on the metrics used by the score function to measure the rationality of triples, these approaches can be classiﬁed into two types: translation models relying on distances and semantic matching models based on semantics.

\subsection{Translation Model}
In the translation model, the scoring function measures the reliability of facts by calculating the distance between two entities.

The most representative model is the Trans series \cite{transe, transd, transr}. Bordes et al. were inspired by the concept of translational invariance and proposed TransE \cite{transe}. This model represents entities and relations in a KG using simple vector forms. Relations between entities are modeled as translation vectors, and it is assumed that the embedding vectors of a valid triple $(h, r, t)$ should satisfy the condition: $h + r \approx t$. The scoring function is defined under the $L_1$ or $L_2$ norm paradigm: $f_r(h,t)=\Vert \mathbf{\mathit{h+r-t}} \Vert_{L_1 / L_2}$. Here, the $L_1$ norm (e.g., Manhattan distance) refers to the sum of the absolute differences between the vector components, while the $L_2$ norm (e.g., Euclidean distance) refers to the square root of the sum of the squared differences between the vector components. These norms are used to measure the closeness between vectors in the embedding space, ensuring that valid triples are scored lower.

Since TransE can only handle one-to-one relations, many extensions have been developed to address one-to-many, many-to-one, and many-to-many relations. For example, Wang et al. proposed TransH \cite{transh}, in which each relation is modeled as a hyperplane. TransR \cite{transr} introduces a relation-specific space, and TransD \cite{transd} constructs a dynamic projection matrix for each entity and relation, simultaneously considering the diversity of entities and relations.

\subsection{Semantic Matching Model}
Unlike the translation model, semantic matching models measure the credibility by the similarity between the underlying semantics of the matched entities and the contained relations in the embedding vector space.

Tensor decomposition is a crucial technique for obtaining low-dimensional vector representations. One notable model in this domain is RESCAL \cite{rescal}, which is based on 3D tensor decomposition. Another model, DistMult \cite{distmult}, uses a neural tensor to learn representations of entities and relations in KGs. To simplify the model, DistMult restricts the relation matrix to a diagonal matrix, which means it can only handle symmetric relations.

Later on, more scholars began to explore the use of neural networks in KRL due to their large parameters and high expressiveness. ConvE \cite{conve}, a 2D convolutional neural network model, emerged as a notable example. ConvE is characterized by its fewer parameters and high computational efficiency. In this model, the interaction between entities and relations is captured by stacking the embedding vectors of head and tail entities into a 2D matrix. Convolutional operations are then applied to this matrix to extract the relationships between them.

The R-GCN \cite{r-gcn} model handles highly multi-relational data features. It assigns different weights to different types of relations, which can easily result in an excessive number of parameters. To address this issue, R-GCN employs two regularization techniques: basis function decomposition and block diagonal decomposition. SACN \cite{sacn} consists of two modules: the encoder WGCN and the decoder Conv-TransE. The WGCN module assigns different weights to different relations, effectively transforming the multi-relation graph into multiple single-relation graphs, each with its own strengths and weaknesses.

However, these pre-LLM KRL methods are designed to preserve the inherent structure of KGs. struggling to effectively represent long-tail entities, as they rely exclusively on the structural information within KGs, favoring entities that are rich in such structural data. Consequently, they fall short in leveraging the textual information embedded in KGs, failing to incorporate the contents of entities and relations into their representations.

\begin{table}[htbp]
  \centering
  \scalebox{0.88}{
    \begin{tabularx}{\textwidth}{m{1.6cm}m{2.6cm}m{3.5cm}m{3.5cm}}
    \toprule
    \textbf{Method}                     & \textbf{Input Type}                           & \textbf{Advantages}                                   & \textbf{Challenges} \\
    \midrule
    Triple \newline Encoding            & Triple as a unit                              & Holistic representation of triples                    & Struggles with unseen triples \\[1.2em]
    
    Translation \newline Encoding       & Head and \newline Relation together           & Improves reasoning by entity-relation dependencies    & Potentially limited expressiveness \\[1.2em]
                                   
    Independent \newline Encoding       & Separate Head, \newline Relation, Tail        & Flexible and modular, supports zero-shot learning     & Lacks integration between elements \\[1.2em]
    \midrule
    Structural \newline Encoding        & Triple sequence                               & Efficient representation of structure and relation    & Dependent on sequence quality \\[1.2em]
    
    Textual \newline Fine-tuning        & Textual triples                               & Simplifies adaptation, reduces need for retraining    & Limited by the generative capabilities \\[1.2em]
    \midrule
    Description \newline Generating     & Triples as \newline prompts                   & Enhances low-resource entity representation           & Heavy reliance on prompt quality \\[1.2em]
    
    Prompt \newline Engineering & Natural language \newline prompts                     & Utilizes vast pre-trained knowledge                   & High computational cost \\[1.2em]
    
    Structural \newline Fine-tuning     & Structural and \newline Textual embeddings    & Combines textual and structural embeddings            & Complex integration of embeddings \\[1.2em]
    \bottomrule
    \end{tabularx}
  }
  \caption{Comparison of Different Methods.}
  \label{tab:method_compare}%
\end{table}%

\section{LLM-Enhanced KRL Methods} \label{Sec_3}
To address the limitations of pre-LLM KRL methods, LLMs have significantly advanced KRL by overcoming the reliance on structural information alone. These models leverage textual information through attention mechanisms, enabling the creation of context-sensitive knowledge representations that better capture the nuances of entities and relations within KGs. This section examines methods to enhance KRL with LLMs, categorized into three approaches: encoder-based (\S \ref{Sec_3.1}), encoder-decoder-based (\S \ref{Sec_3.2}), and decoder-based (\S \ref{Sec_3.3}).

\begin{table}[htbp]
  \centering
  \scalebox{0.88}{
    \begin{tabular}{p{0.5cm} p{2.8cm} p{2.3cm} p{3.5cm} p{1.9cm}}
    \hline
      \textbf{Year} & \textbf{Model} & \textbf{Type}  & \textbf{Base Model} & \textbf{Open Source} \\
    \hline
    \multirow{1}[0]{*}{2019} & KG-BERT \cite{kg-bert} & Encoder & BERT  & \href{https://github.com/yao8839836/kg-bert}{Github} \\
    \hline
    \multirow{3}[0]{*}{2020} & MTL-KGC \cite{mtl-kgc} & Encoder & BERT  & \href{https://github.com/bosung/MTL-KGC}{Github} \\
          & Pretrain-KGE \cite{pretrain-kge} & Encoder & BERT  & - \\
          & K-BERT \cite{k-bert} & Encoder & BERT  & \href{https://github.com/autoliuweijie/K-BERT}{Github} \\
    \hline
    \multirow{9}[0]{*}{2021} & StAR \cite{star} & Encoder & BERT, RoBERTa & \href{https://github.com/wangbo9719/StAR\_KGC}{Github} \\
          & KG-GPT2 \cite{kg-gpt2} & Decoder & GPT2  & - \\
          & BERT-ResNet \cite{bert-resnet} & Encoder & BERT  & \href{https://github.com/justinlovelace/robust-kg-completion}{Github} \\
          & MEM-KGC \cite{mem-kgc} & Encoder & BERT  & - \\
          & LaSS \cite{lass} & Encoder & BERT, RoBERTa & \href{https://github.com/jhshen95/LASS}{Github} \\
          & KEPLER \cite{kepler} & Encoder & RoBERTa & \href{https://github.com/THU-KEG/KEPLER}{Github} \\
          & BLP \cite{blp} & Encoder & BERT  & \href{https://github.com/dfdazac/blp}{Github} \\
          & SimKGC \cite{simkgc} & Encoder & BERT  & \href{https://github.com/intfloat/SimKGC}{Github} \\
          & MLMLM \cite{mlmlm} & Encoder & RoBERTa  & \href{https://github.com/763337092/MLMLM}{Github} \\
    \hline
    \multirow{9}[0]{*}{2022} & LP-BERT \cite{lp-bert} & Encoder & BERT, RoBERTa & - \\
          & PKGC \cite{pkgc} & Encoder & BERT, RoBERTa, LUKE & \href{https://github.com/THU-KEG/PKGC}{Github} \\
          & KGT5 \cite{kgt5} & Encoder-Decoder & T5    & \href{https://github.com/apoorvumang/kgt5}{Github} \\
          & OpenWorld KGC \cite{openworld-kgc} & Encoder & BERT  & - \\
          & $k$NN-KGE \cite{knn-kge} & Encoder & BERT  & \href{https://github.com/zjunlp/KNN-KG}{Github} \\
          & LMKE \cite{lmke} & Encoder & BERT  & \href{https://github.com/Neph0s/LMKE}{Github} \\
          & GenKGC \cite{genkgc} & Encoder-Decoder & BART  & \href{https://github.com/zjunlp/PromptKG/tree/main/research/GenKGC}{Github} \\
          & KG-S2S \cite{kgs2s} & Encoder-Decoder & T5    & \href{https://github.com/chenchens190009/KG-S2S}{Github} \\
    \hline
    \multirow{6}[0]{*}{2023} & LambdaKG \cite{lambdakg} & Encoder-Decoder & BERT, BART, T5 & \href{https://github.com/zjunlp/PromptKG/tree/main/lambdaKG}{Github} \\
          & CSPromp-KG \cite{csprom-kg} & Encoder & BERT  & \href{https://github.com/chenchens190009/CSProm-KG}{Github} \\
          & ReSKGC \cite{reskgc} & Encoder-Decoder & T5    & - \\
          & ReasoningLM \cite{reasoninglm} & Encoder & RoBERTa & \href{https://github.com/RUCAIBox/ReasoningLM}{Github} \\
          & KG-LLM \cite{kg-llm} & Decoder & ChatGLM, LLaMA 2 & \href{https://github.com/yao8839836/kg-llm}{Github} \\
          & KoPA \cite{kopa} & Decoder & Alpaca, LLaMA, GPT 3.5 & \href{https://github.com/zjukg/KoPA}{Github} \\
    \hline
    \multirow{3}[0]{*}{2024} & CD \cite{cd} & Decoder & PaLM2 & \href{https://github.com/DavidLi0406/Contextulization-Distillation}{Github} \\
          & CP-KGC \cite{cp-kgc} & Decoder & Qwen, LLaMA 2, GPT 4 & \href{https://github.com/sjlmg/CP-KGC}{Github} \\
          & KICGPT \cite{kicgpt} & Decoder & GPT 3.5 & - \\
    \hline
    \end{tabular}
  }
  \caption{An overview of various LLM-enhanced KRL models.}
  \label{overview}
\end{table}%

\begin{table}[ht]
\centering
\begin{tabular}{p{2cm} p{10cm}}
\hline
\textbf{Notation} & \textbf{Description} \\
\hline
$h$, $r$, $t$ & Head entity, Relation, Tail entity \\
$\mathbf{X}{(h, r)}$, $\mathbf{X}(t)$ & Input sequences for head-relation and tail entity \\
$D$, $D^+$, $D^-$ & Set of all triples, positive triples, negative triples \\
$\mathbf{u}$, $\mathbf{v}$ & Contextualized embeddings of head-relation pair and tail entity \\
$s_{c}$, $s_{d}$ & Classification score, Distance score \\
$\mathbf{h}$, $\mathbf{r}$, $\mathbf{t}$ & Embeddings for head entity, relation, and tail entity \\
$L_c$, $L_d$, $L$ & Classification loss, Distance-based loss, Total loss \\
$\lambda$, $\gamma$, $n$, $K$ & Margin, Weight factor, Number of negative samples, Sum limit \\
$\sigma$ & Sigmoid function \\
$\alpha$, $\beta$ & Auxiliary loss weights \\
$\mathbf{K}$, $I$, $X$ & Transformed structural embeddings, Instruction prompt, Triple prompt \\
\hline
\end{tabular}
\caption{Notations of Concepts.}
\label{notation_table}
\end{table}

Table \ref{tab:method_compare} presents a comparison of various methods utilized for KRL with LLMs, emphasizing their input types, advantages, and challenges. Each method adopts a unique approach to encoding and processing triples or related information, catering to specific tasks and scenarios within the broader scope of KRL. Encoder-based approaches like triple-based representation treats the triple as a single unit, providing a holistic representation but struggling with unseen triples. Translation-based representation pairs the head and relation for improved reasoning, although it may be less expressive. Independent representation handles head, relation, and tail separately, supporting modularity and zero-shot learning but lacking integration across elements. Similarly, encoder-decoder-based structure-based representation methods leverage triple sequences for efficient structural representation, while textual fine-tuning simplifies adaptation by focusing on textual triples but is limited by the generative power of the underlying model. Decoder-based approaches like description generation, which enhances representation in low-resource settings via prompts but is heavily dependent on prompt quality, and prompt engineering, which harnesses the pre-trained knowledge of LLMs but at high computational costs. Lastly, structural fine-tuning combines structural and textual embeddings to balance textual and structural knowledge, though it requires sophisticated integration.

An overview of various LLM-enhanced KRL models is provided in Table \ref{overview}, which provides a comprehensive summary of various LLM-enhanced KRL models, organized by year, model name, type, base model, and open-source availability. This table highlights key trends in the field, such as the early dominance of encoder-based architectures leveraging BERT \cite{bert} (e.g., KG-BERT \cite{kg-bert}, StAR \cite{star}) and the growing adoption of encoder-decoder models (e.g., GenKGC \cite{genkgc}, KGT5 \cite{kgt5}) in recent years for tasks like generative knowledge graph completion. The increasing use of decoder-based LLMs (e.g., KG-LLM \cite{kg-llm}, KoPA \cite{kopa}) in 2023 and 2024 indicates a shift towards leveraging generative and reasoning capabilities. Open-source contributions from many models (e.g., KEPLER \cite{kepler}, PKGC \cite{pkgc}) have played a crucial role in advancing the field, though some models remain proprietary. 

To establish a unified framework for the symbols and formulas utilized in subsequent sections, Table \ref{notation_table} summarizes the main notations and their descriptions.

\subsection{Encoder-Based Methods} \label{Sec_3.1}
Encoder-based methods mainly employ encoders like BERT \cite{bert} and RoBERTa \cite{roberta}, offering the advantage of leveraging abundant textual information, such as entity descriptions, to enhance KRL. These methods differ in their encoding of KG triples. In triple-based representation (\S \ref{Sec_3.1.1}), the head entity (HE), relation, and tail entity (TE) are encoded as a single unit. Conversely, translation-based representation (\S \ref{Sec_3.1.2}) encodes the head entity and relation together, with the tail entity encoded separately, optimizing through vector space distances. independent representation (\S \ref{Sec_3.1.3}) processes the head entity, relation, and tail entity as separate inputs, encoding each independently.

\subsubsection{Triple-based Representation} \label{Sec_3.1.1}
\begin{figure*}
\centering
\includegraphics[width=0.6\textwidth]{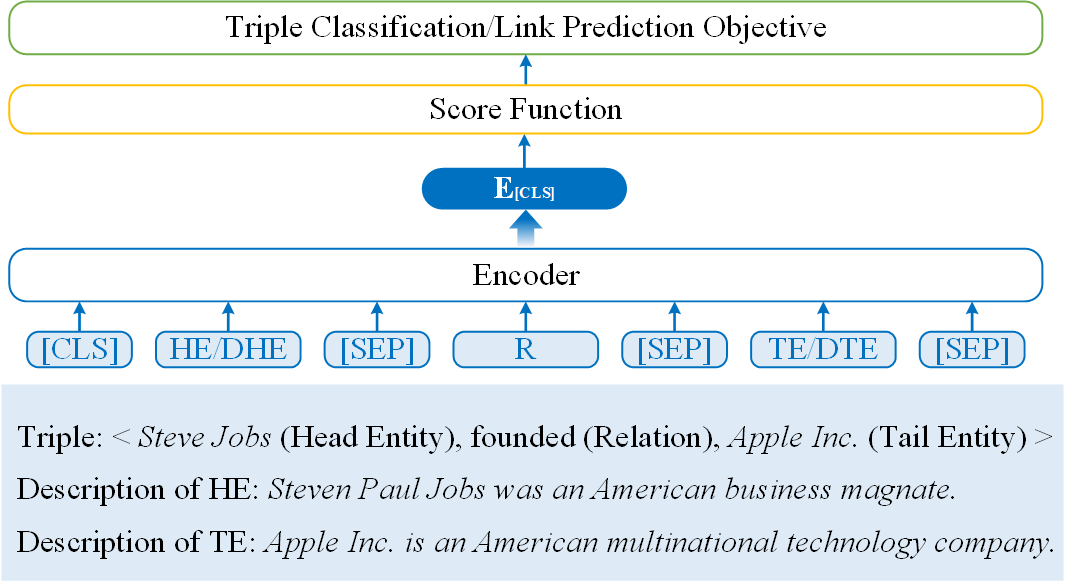}
\caption{An overview of triple-based representation methods.}
\label{Fig_3.1.1}
\end{figure*}

Triple-based representation methods represent the entire knowledge graph triple as a single unit, leveraging the integrated semantic and relational information of the head and tail entities as well as the relation. This holistic representation not only enhances downstream tasks but also captures the contextual significance of entities by fully utilizing rich language patterns. Fig. \ref{Fig_3.1.1} illustrates a typical model, KG-BERT \cite{kg-bert}, which uses triple-based representation.

KG-BERT treats triples as textual sequences and fine-tunes a pre-trained BERT model to predict the plausibility of a given triple. The model takes as input the descriptions of the head entity (DHE) $h$, relation (DR) $r$, and tail entity (DTE) $t$:

\begin{equation}
\text{Input sequence: } [\text{CLS}] \; \text{Tok}_{hi} \; [\text{SEP}] \; \text{Tok}_{ri} \; [\text{SEP}] \; \text{Tok}_{ti} \; [\text{SEP}]
\end{equation}

\noindent where $\text{Tok}_{hi}$, $\text{Tok}_{ri}$, and $\text{Tok}_{ti}$ represent the tokens of the head entity, relation, and tail entity descriptions, respectively. The model computes the final hidden vector of the unique [CLS] token, which serves as the aggregate sequence representation. The plausibility score of a triple $\tau = (h, r, t)$ is then computed as follows:

\begin{equation}
s_\tau = \text{sigmoid}(C W^T)
\end{equation}

\noindent where $C$ is the aggregated sequence representation derived from the [CLS] token, KG-BERT emphasizes semantic nuances and relational integrity even when triples are incomplete or corrupted. The cross-entropy-based loss function facilitates robust learning by effectively differentiating between noisy negative samples and valid triples:

\begin{equation}
L = -\sum_{\tau \in D^+ \cup D^-} (y_\tau \log(s_{\tau0}) + (1 - y_\tau) \log(s_{\tau1}))
\end{equation}

\noindent where $y_\tau \in \{0, 1\}$ is the label indicating whether the triple is positive or negative, and $D^+$ and $D^-$ are the sets of positive and negative triples, respectively. Negative triples are generated by corrupting either the head or the tail entity of a positive triple. Negative triples, generated through entity corruption, simulate real-world ambiguities, making the training process more resilient to such scenarios.

MTL-KGC \cite{mtl-kgc} employs a multi-task learning framework to enhance knowledge graph completion. It incorporates additional tasks like relation prediction and relevance ranking, significantly improving upon models such as KG-BERT by learning more relational properties and effectively distinguishing between lexically similar candidates.

K-BERT \cite{k-bert} integrates structured domain knowledge from knowledge graphs directly into the pre-training process. It retains the original BERT architecture but adds a mechanism to inject relevant triples from a knowledge graph into the training examples, enriching the language model with domain-specific information.

MLMLM \cite{mlmlm} introduces a novel method for link prediction using Mean Likelihood Masked Language Models to generate potential entities directly. This approach leverages the knowledge embedded in pre-trained MLMs to enhance the interpretability and scalability of knowledge graphs.

PKGC \cite{pkgc} is a pre-trained language model (PLM) based model for knowledge graph completion that utilizes prompt engineering to enhance knowledge utilization. This model adapts PLM for the task by converting knowledge triples into natural language prompts, facilitating the effective use of latent knowledge of the PLM.

CSProm-KG \cite{csprom-kg} incorporates both structural and textual information into the knowledge graph completion process. This model employs Conditional Soft Prompts that adapt the inputs based on the structure of the knowledge graph, allowing pre-trained language models to efficiently utilize both types of information.

\subsubsection{Translation-based Representation} \label{Sec_3.1.2}
\begin{figure*}
\centering
\includegraphics[width=0.6\textwidth]{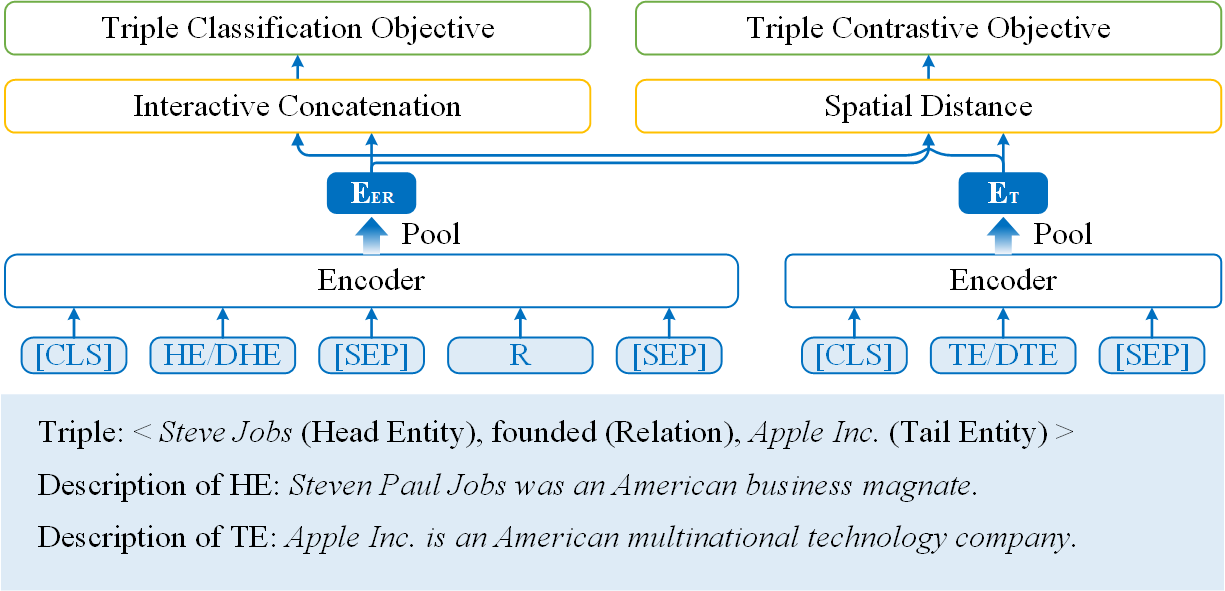}
\caption{An overview of translation encoding methods.}
\label{Fig_3.1.2}
\end{figure*}

Translation-based representation methods focus on encoding the head entity and relation while treating the tail entity separately. This technique relies on optimizing the vector space distances between these entities, facilitating more accurate knowledge representation and reasoning. Fig. \ref{Fig_3.1.2} depicts a representative model, StAR \cite{star}, that employs translation-based representation.

This model divides each triple into two asymmetric parts: one combining the head entity and relation, and the other consisting of the tail entity. The StAR model employs a Siamese-style textual encoder to convert these parts into contextualized representations. The translation function is defined as follows:

\begin{equation}
\mathbf{u} = \text{Pool}(\text{Transformer-Enc}(\mathbf{X}_{(h, r)})), \quad \mathbf{v} = \text{Pool}(\text{Transformer-Enc}(\mathbf{X}_{t}))
\end{equation}

\noindent where $\mathbf{X}_{(h, r)} = [\text{CLS}] \; \mathbf{X}_h \; [\text{SEP}] \; \mathbf{X}_r \; [\text{SEP}]$ and $\mathbf{X}_{t} = [\text{CLS}] \; \mathbf{X}_t \; [\text{SEP}]$ are the input sequences for the head-relation pair and the tail entity, respectively. $\text{Pool}(\cdot)$ denotes the pooling operation to obtain the sequence-level representation from the transformer's output.

The plausibility score of a triple $(h, r, t)$ is calculated using a combination of a deterministic classifier and a spatial measurement:

\begin{equation}
s_{c} = \text{softmax}(\text{MLP}([\mathbf{u}; \mathbf{u} \circ \mathbf{v}; \mathbf{u} - \mathbf{v}; \mathbf{v}])), \quad s_{d} = -\|\mathbf{u} - \mathbf{v}\|_2
\end{equation}

\noindent where $[\mathbf{u}; \mathbf{u} \circ \mathbf{v}; \mathbf{u} - \mathbf{v}; \mathbf{v}]$ denotes the interactive concatenation of the contextualized embeddings $\mathbf{u}$ and $\mathbf{v}$, and $\text{MLP}(\cdot)$ is a multi-layer perceptron.

The model is trained using both a triple classification objective and a contrastive loss:

\begin{equation}
L_c = -\frac{1}{|D|} \sum_{t_p \in D} \left( \log s_{c} + \sum_{t_p' \in N(t_p)} \log(1 - s_{c}') \right)
\end{equation}

\begin{equation}
L_d = \frac{1}{|D|} \sum_{t_p \in D} \sum_{t_p' \in N(t_p)} \max(0, \lambda - s_{d} + s_{d}')
\end{equation}

\noindent where $D$ is the set of positive triples, $N(t_p)$ denotes the set of negative triples generated by corrupting positive triples, and $\lambda$ is the margin.

The overall training loss is a weighted sum of the two objectives:

\begin{equation}
L = L_c + \gamma L_d
\end{equation}

\noindent where $\gamma$ is a hyperparameter that balances the two losses.

SimKGC \cite{simkgc} uses a bi-encoder architecture to enhance contrastive learning efficiency in knowledge graph completion. It incorporates three types of negative sampling—in-batch, pre-batch, and self-negatives—improving training efficacy through larger negative sets and a novel InfoNCE loss function.

LP-BERT \cite{lp-bert} employs a multi-task pre-training strategy, enhancing knowledge graph completion by predicting relations between entities. This method integrates masked language, entity, and relation modeling tasks, significantly improving link prediction. It effectively manages unseen entities and relations by leveraging contextualized entity and relation information from large datasets.

\begin{figure*}
\centering
\includegraphics[width=0.6\textwidth]{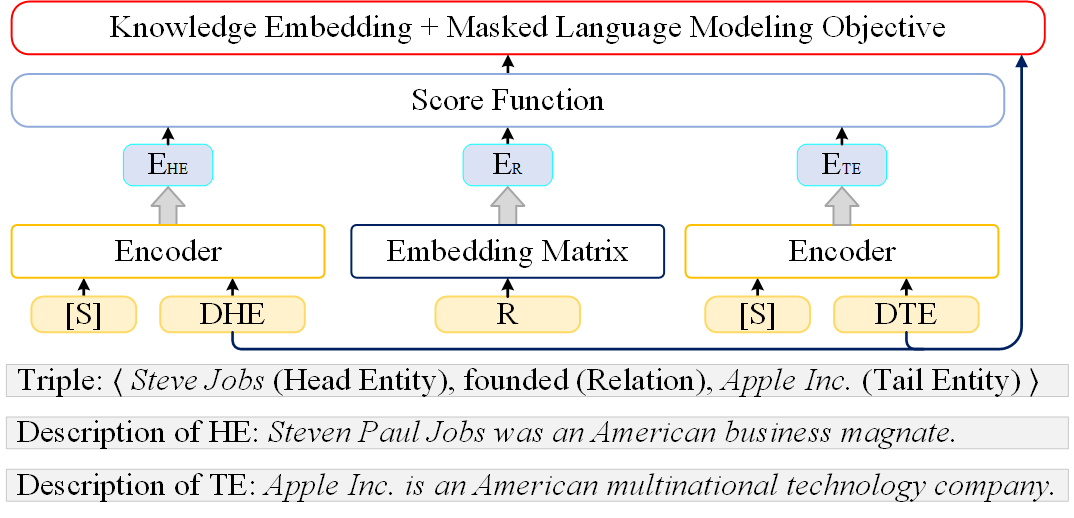}
\caption{An overview of independent representation methods.}
\label{Fig_3.1.3}
\end{figure*}

\subsubsection{Independent Representation} \label{Sec_3.1.3}
Independent representation methods separately encode each component of a triple: the head entity, relation, and tail entity. This approach allows for flexible and modular representations that benefit specific knowledge graph applications. A significant advantage of independent representation is its ability to enhance structural information within the KG, providing zero-shot capabilities. Fig. \ref{Fig_3.1.3} shows the architecture of a typical independent representation model, KEPLER \cite{kepler}.

KEPLER generates embeddings using textual descriptions of entities, which reduces reliance on the frequency of training samples. This approach ensures that even low-frequency entities can achieve high-quality representations through their textual descriptions, mitigating degradation in embedding quality caused by imbalanced data:

\begin{equation}
h = E_{\langle s \rangle}(\text{text}_h), \quad t = E_{\langle s \rangle}(\text{text}_t), \quad r = T_r,
\end{equation}

\noindent where $\text{text}_h$ and $\text{text}_t$ are the descriptions for the head and tail entities, respectively, each prefixed with a special token $\langle s \rangle$. $T \in \mathbb{R}^{|R| \times d}$ is the relation embeddings matrix, and $h$, $t$, and $r$ represent the embeddings for the head entity, tail entity, and relation, respectively. The encoding function $E_{\langle s \rangle}$ leverages pre-trained language models to capture contextual and semantic relationships instead of depending solely on entity frequency.

KEPLER's knowledge embedding objective uses negative sampling for optimization:

\begin{equation}
L_{KE} = - \log \sigma(\gamma - d_r(h, t)) - \frac{1}{n} \sum_{i=1}^{n} \log \sigma(d_r(h'_i, t'_i) - \gamma),
\end{equation}

\noindent where $(h'_i, r, t'_i)$ are negative samples, $\gamma$ is the margin, $\sigma$ is the sigmoid function. The described mechanism ensures that embeddings are derived from descriptive semantics rather than raw frequency, mitigating the risk of over-representing frequently occurring entities while under-representing rare ones. $d_r$ is the scoring function:

\begin{equation}
d_r(h, t) = \|h + r - t\|_p,
\end{equation}

\noindent with the norm $p$ set to 1. The negative sampling policy involves fixing the head entity and randomly sampling a tail entity, and vice versa.

The BERT-ResNet \cite{bert-resnet} model extends the capabilities of encoder-based KRL methods by integrating BERT with a deep residual network. This combination enhances the handling of sparse connectivity in knowledge graphs. The model leverages BERT's robust embeddings and ResNet's deep convolutional architecture, significantly improving entity ranking performance even with limited training data.

The BLP \cite{blp} model focuses on inductive link prediction by utilizing textual entity descriptions and pre-trained language models, emphasizing its capacity to generalize across unseen entities. This model surpasses previous approaches by incorporating dynamic graph embeddings, allowing it to effectively adapt to continuously evolving knowledge graphs without retraining.

\subsection{Encoder-Decoder-Based Methods} \label{Sec_3.2}
Encoder-decoder-based methods employ models such as BART \cite{bart} and T5 \cite{t5}, are known for their intuitive simplicity, as all desired functionalities can be achieved through a straightforward seq2seq model. These methods are classified by the type of input sequence utilized. Structure-based representation methods (\S \ref{Sec_3.2.1}) resemble Encoder-based methods, utilizing encoder-type triple sequences as input. Fine-tuning methods (\S \ref{Sec_3.2.2}), similar to Decoder-based methods, use natural language expressions of triples as decoder input.

\subsubsection{Structure-based Representation} \label{Sec_3.2.1}
\begin{figure*}
\centering
\includegraphics[width=0.6\textwidth]{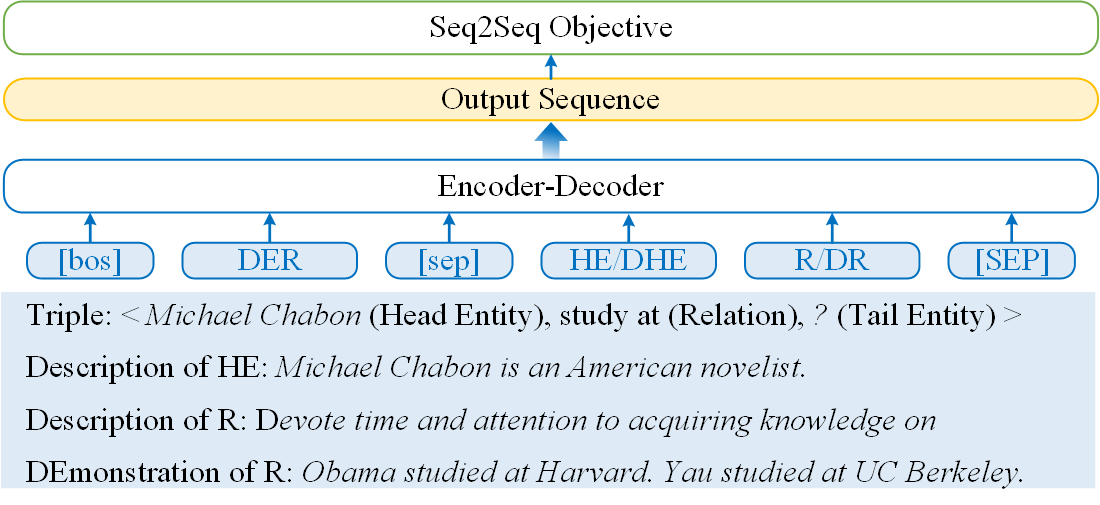}
\caption{An overview of structure-based representation methods.}
\label{Fig_3.2.1}
\end{figure*}

Structure-based representation leverages the structure of the triples by feeding them into the encoder as sequences. which allows the model to capture both the syntactic and relational structure of the triples. By integrating structural information with natural language input, this method enhances the ability to represent complex relations. Fig. \ref{Fig_3.2.1} demonstrates the GenKGC \cite{genkgc} architecture that exemplifies structure-based representation.

This approach transforms knowledge graph completion into a sequence-to-sequence generation task, leveraging pre-trained language models to generate target entities from input sequences representing head entities and relations. The structure-based representation is as follows:

Given a triple with a missing tail entity $(e_h, r, ?)$, the input sequence is constructed by concatenating the descriptions of the head entity $d_{e_h}$ and the relation $d_r$:

\begin{equation}
\text{Input sequence: } [\text{CLS}] \; d_{e_h} \; [\text{SEP}] \; d_r \; [\text{SEP}]
\end{equation}

\noindent where $d_{e_h}$ and $d_r$ are the textual descriptions of the head entity and the relation, respectively. The output sequence corresponds to the target tail entity $d_{e_t}$.

The generative process is defined as:

\begin{equation}
p_{\theta}(y|x) = \prod_{i=1}^{N} p_{\theta}(y_i | y_{<i}, x)
\end{equation}

\noindent where $x$ is the input sequence, $y$ is the output sequence, and $N$ is the length of the output sequence. The model is trained using a standard sequence-to-sequence objective function:

\begin{equation}
L = -\log p_{\theta}(y|x)
\end{equation}

GenKGC improves representation learning and reduces inference time through relation-guided demonstration and entity-aware hierarchical decoding. Relation-guided demonstration augments the input sequence with examples of triples that share the same relation:

\begin{equation}
x = [\text{CLS}] \; \text{demonstration}(r) \; [\text{SEP}] \; d_{e_h} \; [\text{SEP}] \; d_r \; [\text{SEP}]
\end{equation}

Entity-aware hierarchical decoding constrains the decoding process using entity-type information, which reduces the search space during generation and enhances inference efficiency.

LambdaKG \cite{lambdakg} advances the application of pre-trained language models in knowledge graph embeddings by integrating structural and textual information into a unified model architecture. It employs advanced training techniques such as prompt engineering, negative sampling, and relation-aware modeling to enhance the efficiency and accuracy of knowledge graph representations.

\subsubsection{Textual Fine-Tuning} \label{Sec_3.2.2}
\begin{figure*}
\centering
\includegraphics[width=0.6\textwidth]{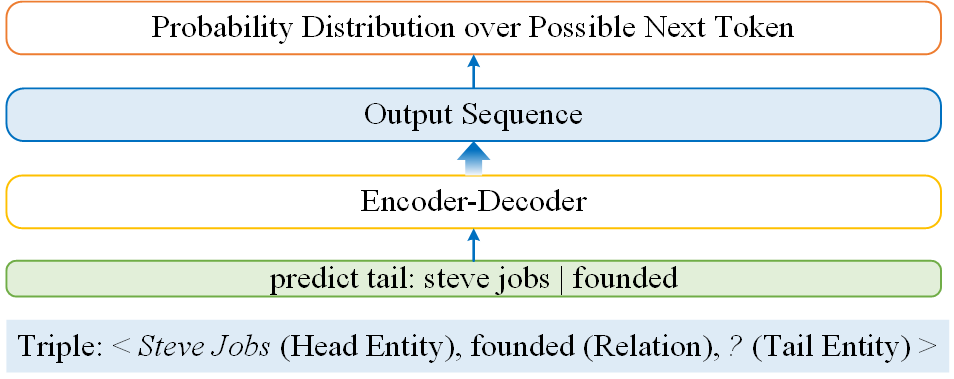}
\caption{An overview of fine-tuning methods.}
\label{Fig_3.2.2}
\end{figure*}

Textual fine-tuning methods adapt pre-trained encoder-decoder models using natural language descriptions of triples. This approach tailors the model's generative capabilities to specific knowledge representation tasks, offering significant advantages in terms of training and inference efficiency. An illustration of a typical textual fine-tuning model KGT5 \cite{kgt5} is presented in Fig. \ref{Fig_3.2.2}.

KGT5 fine-tunes a T5 model for both link prediction and question answering (QA) tasks. This process involves initial pre-training on link prediction, followed by fine-tuning on QA datasets, balancing these tasks using a regularization approach.

Given a set of entities $E$ and relations $R$, a knowledge graph $K \subseteq E \times R \times E$ consists of triples $(s, p, o)$. KGT5 treats KG link prediction and QA as sequence-to-sequence tasks, utilizing textual representations of entities and relations. A verbalization scheme is employed to convert link prediction queries into textual queries.

The training process is described by the following formulas:

\begin{equation}
\text{Score}(a) = p_a + \alpha \quad \text{if} \; a \in N(e)
\end{equation}

\begin{equation}
\text{Score}(a) = p_a \quad \text{otherwise}
\end{equation}

\noindent where $p_a$ is the log probability of the predicted entity $a$, $\alpha$ is a constant hyperparameter, and $N(e)$ represents the $n$-hop neighborhood of the entity $e$.

The model's architecture and training process effectively reduce the model size while maintaining or improving performance on large KG and KGQA tasks.

During the fine-tuning phase for QA tasks, KGT5 continues training on link prediction objectives. This dual-task regularization ensures that the model retains its ability to generalize beyond the specific QA dataset, as the link prediction tasks anchor the learning to broader KG. Each training batch contains an equal number of QA examples and link prediction examples. This balanced batching prevents overfitting to the QA data while ensuring that the model remains grounded in the broader structure of the KG.

KG-S2S \cite{kgs2s} is a state-of-the-art sequence-to-sequence (Seq2Seq) generative framework designed to address various challenges in knowledge graph completion (KGC) across different settings without requiring structural modifications for each graph type. This model uniquely overcomes the limitations of previous KGC approaches that are tightly coupled with specific graph structures, restricting their adaptability to new or evolving KGs. By treating all elements of knowledge graphs—entities, relations, and metadata—as sequences in a unified ‘flat’ text format, KG-S2S simplifies data representation and enhances the model’s flexibility and scalability. It employs advanced fine-tuning techniques on pre-trained language models, integrating novel mechanisms such as entity descriptions and relational soft prompts to enrich the model’s contextual understanding.

\subsection{Decoder-Based Methods} \label{Sec_3.3}
Decoder-based methods, utilizing models like LLaMA \cite{llama2} and GPT-4 \cite{gpt4}, are defined by the critical function of the decoder in the representation learning process. These methods leverage much larger semantic knowledge without incurring additional training overhead. Description generation methods (\S \ref{Sec_3.3.1}) produce descriptive text for low-resource entities, employing encoder-based or encoder-decoder-based methods for other tasks. Prompt engineering (\S \ref{Sec_3.3.2}) utilizes the decoder as a question-answering tool, leveraging natural language to retrieve triple information and execute downstream tasks. Structural fine-tuning (\S \ref{Sec_3.3.3}) integrates structural and textual embeddings, refining the decoder's output to enhance knowledge representation.

\subsubsection{Description Generation} \label{Sec_3.3.1}
\begin{figure*}
\centering
\includegraphics[width=0.6\textwidth]{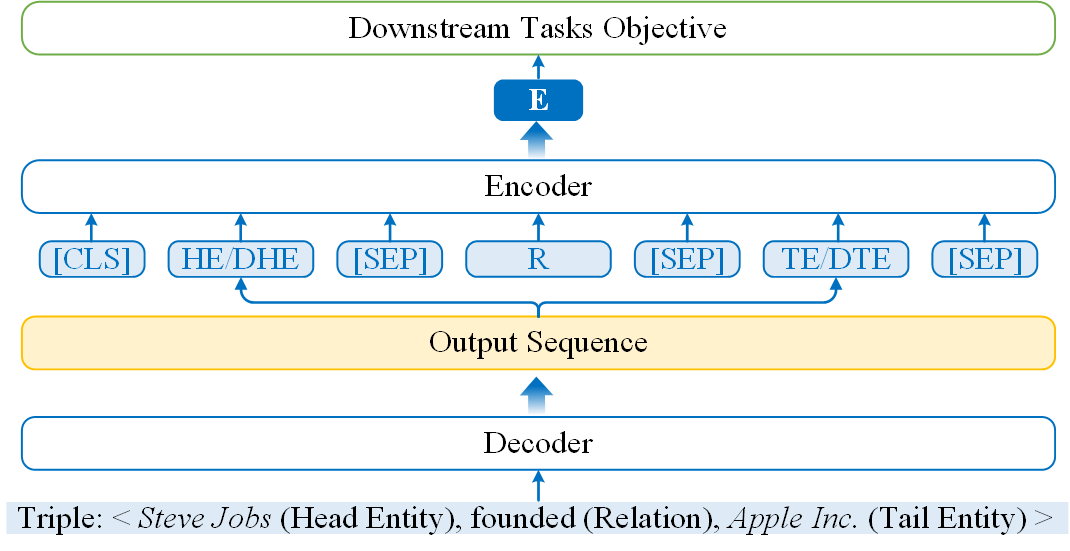}
\caption{An overview of description generation methods.}
\label{Fig_3.3.1}
\end{figure*}

Description generation methods enhance the representation of low-resource entities by creating descriptive text. This approach supplements the textual information, improving the performance of previous methods and ensuring a richer, more complete entity representation. Fig. \ref{Fig_3.3.1} illustrates a typical contextualization distillation (CD) \cite{cd} method used in description generation method.

This method utilizes LLMs to convert concise structural triplets into rich, contextual segments. The process involves generating descriptive contexts for knowledge graph completion using prompts. The main steps are as follows:

Given a triplet $(h, r, t)$ where $h$ is the head entity, $r$ is the relation, and $t$ is the tail entity, the method generates descriptive context $c$ using an LLM prompted with the triplet. The process can be formulated as:

\begin{equation}
p_i = \text{Template}(h_i, r_i, t_i)
\end{equation}

\begin{equation}
c_i = \text{LLM}(p_i)
\end{equation}

\noindent where $p_i$ is the prompt template filled with the triplet components, and $c_i$ is the descriptive context generated by the LLM.

Two auxiliary tasks, reconstruction and contextualization, are introduced to train smaller knowledge graph completion (KGC) models with these enriched triplets. The reconstruction task aims to restore corrupted descriptive contexts using masked language modeling (MLM), while the contextualization task trains the model to generate descriptive context from the original triplet. The losses for these tasks are defined as:

\begin{equation}
L_{\text{rec}} = \frac{1}{N} \sum_{i=1}^{N} \ell(f(\text{MLM}(c_i)), c_i)
\end{equation}

\begin{equation}
L_{\text{con}} = \frac{1}{N} \sum_{i=1}^{N} \ell(f(I_i), c_i)
\end{equation}

\noindent where $\ell$ is the cross-entropy loss function, $f$ is the model, $I_i$ is the concatenated input of the head, relation, and tail, and $N$ is the number of samples.

The final loss for training the KGC models combines the KGC loss with the auxiliary task losses:

\begin{equation}
L_{\text{final}} = L_{\text{kgc}} + \alpha \cdot L_{\text{rec}} + \beta \cdot L_{\text{con}}
\end{equation}

\noindent where $\alpha$ and $\beta$ are hyperparameters that balance the contributions of the auxiliary losses.

The CP-KGC \cite{cp-kgc} model represents a significant advancement in text-based KGC. It specifically leverages Constrained-Prompt Knowledge Graph Completion with Large Language Models to refine and enhance textual descriptions within KGC datasets. CP-KGC uses simple, carefully designed prompts to regenerate or supplement existing textual descriptions, improving the overall expressiveness and utility of the data. This approach effectively addresses issues like hallucinations in text generation by LLMs, ensuring more accurate and contextually relevant outputs.

\subsubsection{Prompt Engineering} \label{Sec_3.3.2}
\begin{figure*}
\centering
\includegraphics[width=0.6\textwidth]{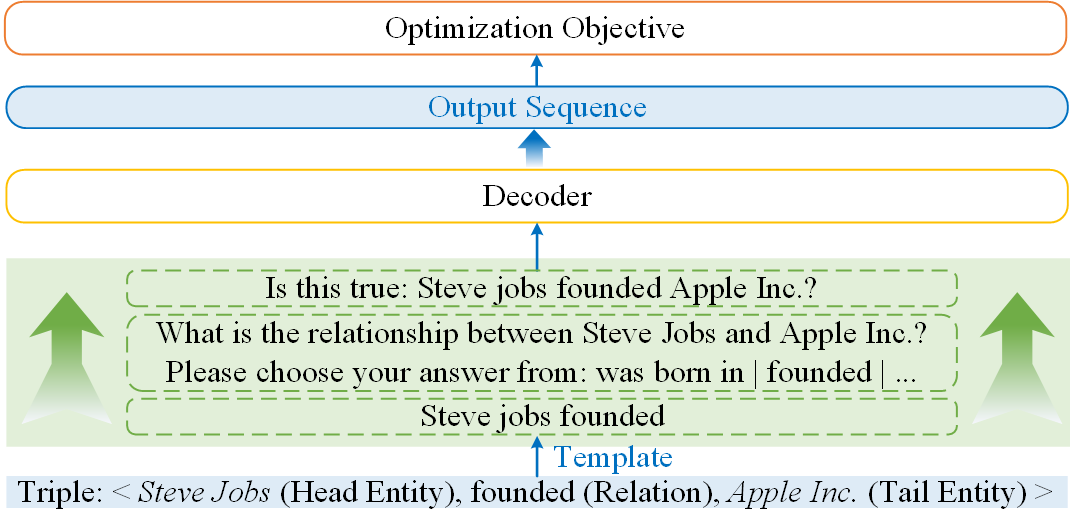}
\caption{An overview of prompt engineering methods.}
\label{Fig_3.3.2}
\end{figure*}

Prompt engineering leverages the natural language capabilities of decoders, framing knowledge retrieval and representation tasks as question-answering problems, which utilizes the vast amount of real-world knowledge within the model. Fig. \ref{Fig_3.3.2} shows a representative model KG-LLM \cite{kg-llm} using prompt engineering.

This method transforms triples into natural language prompts and employs LLMs to predict the plausibility of the triples or complete missing information. The process is as follows:

For a given triple $(h, r, t)$, where $h$ is the head entity, $r$ is the relation, and $t$ is the tail entity, a prompt is constructed to query the LLM. For example, for triple classification, the prompt might be:

\begin{equation}
\texttt{Prompt: Is it true that } h \; r \; t\text{?}
\end{equation}

The LLM then generates a response indicating the plausibility of the triple. A scoring function quantifies this plausibility:

\begin{equation}
s(h, r, t) = \text{LLM}(p(h, r, t))
\end{equation}

\noindent where $p(h, r, t)$ is the prompt generated for the triple $(h, r, t)$, and $\text{LLM}$ represents the large language model.

The training objective uses a cross-entropy loss to fine-tune the LLM on labeled triples:

\begin{equation}
L = - \sum_{(h,r,t) \in D} [y \log s(h, r, t) + (1 - y) \log (1 - s(h, r, t))]
\end{equation}

\noindent where $D$ is the set of training triples, and $y$ is the label indicating the truth of the triple.

The study performs specific instruction tuning on LLMs, aligning their general natural language understanding abilities with the processing of knowledge graph triples. Fine-tuned models (e.g., KG-LLaMA and KG-ChatGLM) extract and utilize knowledge representations more efficiently, further minimizing errors caused by language ambiguity.

The KICGPT \cite{kicgpt} model integrates an LLM with knowledge graph completion (KGC) methods to address challenges in traditional KGC approaches. It significantly reduces training overhead by using in-context learning strategies, eliminating the need for explicit model fine-tuning. The model leverages the LLM’s extensive pre-trained knowledge base and a structure-aware KG retriever to improve the handling of long-tail entities. This integration allows KICGPT to effectively utilize both structured KG information and the broad knowledge base of the LLM, providing a more robust framework for KGC tasks. Notably, the model employs ‘Knowledge Prompt’ strategies, guiding the LLM with structured prompts that incorporate KG information, thus enhancing its ability to make informed predictions about missing entities in the KG.

\subsubsection{Structural Fine-Tuning} \label{Sec_3.3.3}
\begin{figure*}
\centering
\includegraphics[width=0.6\textwidth]{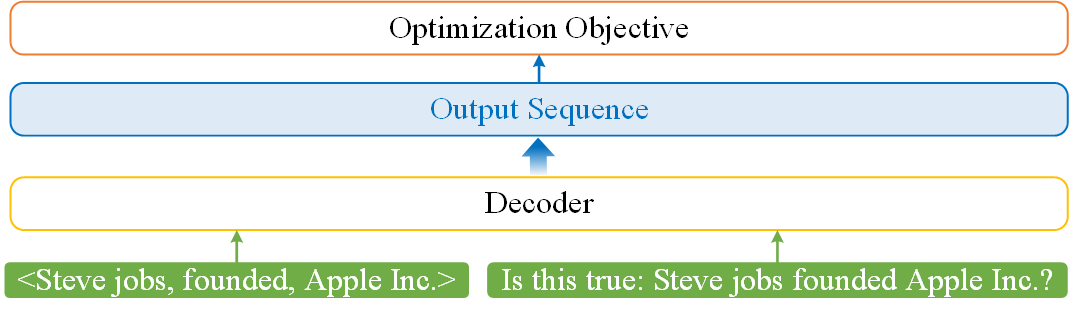}
\caption{An overview of structural fine-tuning methods.}
\label{Fig_3.3.3}
\end{figure*}

Structural fine-tuning combines embeddings from structural and textual information, feeding them into the decoder. By incorporating structural information from KGs, this approach optimizes the output of LLMs, providing a more comprehensive representation of the KG triples. Fig. \ref{Fig_3.3.3} depicts a structural fine-tuning model KoPA \cite{kopa}.

This method incorporates structural embeddings of entities and relations into LLMs to improve knowledge graph completion tasks.

KoPA's core concept involves two primary steps: pre-training structural embeddings and using a prefix adapter to inject these embeddings into the LLM. The structural embeddings are first learned through a self-supervised pre-training process, capturing the knowledge graph's structural information. The process is as follows:

For each triple $(h, r, t)$, the structural embeddings $ \mathbf{h}, \mathbf{r}, \mathbf{t} $ are learned using the scoring function:

\begin{equation}
F(h, r, t) = \|\mathbf{h} + \mathbf{r} - \mathbf{t}\|^2
\end{equation}

The pre-training objective involves a margin-based ranking loss with negative sampling:

\begin{equation}
L_{\text{pre}} = -\log \sigma(\gamma - F(h, r, t)) - \sum_{i=1}^{K} \log \sigma(F(h'_i, r'_i, t'_i) - \gamma)
\end{equation}

\noindent where $(h'_i, r'_i, t'_i)$ are negative samples, $ \gamma $ is the margin, and $ \sigma $ is the sigmoid function.

Once pre-trained, KoPA uses a prefix adapter to map these structural embeddings into the LLM's textual token space. This transformation is performed as follows:

\begin{equation}
\mathbf{K} = P(\mathbf{h}) \oplus P(\mathbf{r}) \oplus P(\mathbf{t})
\end{equation}

\noindent where $P$ is a projection layer, and $\oplus$ denotes concatenation. The transformed embeddings, known as virtual knowledge tokens, are prepended to the input sequence $S$:

\begin{equation}
S_{\text{KoPA}} = \mathbf{K} \oplus I \oplus X
\end{equation}

\noindent where $I$ is the instruction prompt, and $X$ is the triple prompt.

The fine-tuning objective for the LLM with KoPA involves minimizing the cross-entropy loss:

\begin{equation}
L_{\text{KoPA}} = -\frac{1}{|S_{\text{KoPA}}|} \sum_{i=1}^{|S_{\text{KoPA}}|} \log P_{\text{LLM}}(s_i | s_{<i})
\end{equation}

\noindent where $s_i$ are the tokens in the input sequence $S_{\text{KoPA}}$.

KG-GPT2 \cite{kg-gpt2} is an adaptation of the GPT-2 language model for knowledge graph completion. KG-GPT2 leverages GPT-2's contextual capabilities to address incomplete knowledge graphs by predicting missing links and relationships. This model treats each triple in a knowledge graph as a sentence, allowing GPT-2 to accurately classify the likelihood of triples. By contextualizing triples, KG-GPT2 surpasses traditional embedding techniques, incorporating richer linguistic and semantic information to enhance graph completion.

\section{Experiments and Evaluations} \label{Sec_4}
This section systematically reviews existing experiments and evaluations to analyze the performance and effectiveness of various LLM-enhanced KRL methods discussed previously. By consolidating results from various studies, we aim to present a structured overview of their potential and constraints. The best and second-best results in tables are shown in bold and underlined, respectively.

\begin{table}[htbp]
  \centering
  \scalebox{0.88}{
    \begin{tabularx}{\textwidth}{p{2.3cm} XX p{1.3cm} XXXXX}
    \hline
    \textbf{Dataset} & \textbf{\# Ent} & \textbf{\# Rel} & \textbf{\# Train} & \textbf{\# Dev} & \textbf{\# Test} & \textbf{Task}  & \textbf{Year} \\
    \hline
    FIGER \cite{figer} & -     & -     & 2,000,000 & 10,000 & 563   & ET    & 2015 \\
    Open Entity \cite{open_entity} & -     & -     & 2,000 & 2,000 & 2,000 & ET    & 2018 \\
    \hline
    TACRED \cite{tacred} & -     & 42    & 68,124 & 22,631 & 15,509 & RC    & 2017 \\
    FewRel \cite{fewrel} & -     & 80    & 8,000 & 16,000 & 16,000 & RC    & 2018 \\
    \hline
    WN11 \cite{wn11_fb13} & 38,696 & 11    & 112,581 & 2,609 & 10,544 & TC    & 2013 \\
    FB13 \cite{wn11_fb13} & 75,043 & 13    & 316,232 & 5,908 & 23,733 & TC    & 2013 \\
    WN9 \cite{wn9}  & 6,555 & 9     & 11,741 & 1,337 & 1,319 & TC    & 2022 \\
    FB15K-237N \cite{fb15k-237n} & 13,104 & 93    & 87,282 & 7,041 & 8,226 & TC    & 2022 \\
    \hline
    Nations \cite{nations} & 14    & 55    & 1,592 & 199   & 201   & LP    & 2007 \\
    FB15K \cite{fb15k} & 14,951 & 1,345 & 483,142 & 50,000 & 59,071 & LP    & 2013 \\
    FB15k-237 \cite{fb15k-237} & 14,541 & 237   & 272,115 & 17,535 & 20,466 & LP    & 2015 \\
    WN18RR \cite{wn18rr_umls_yago3-10} & 40,943 & 11    & 86,835 & 3,034 & 3,134 & LP, LP(ZS) & 2018 \\
    UMLS \cite{wn18rr_umls_yago3-10} & 135   & 46    & 5,216 & 652   & 661   & LP    & 2018 \\
    YAGO3-10 \cite{wn18rr_umls_yago3-10} & 103,222 & 30    & 490,214 & 2,295 & 2,292 & LP    & 2018 \\
    NELL-ONE \cite{nell-one} & 68,545 & 822   & 189,635 & 1,004 & 2,158 & LP(ZS) & 2018 \\
    CoDEx-S \cite{codex} & 45,869 & 68    & 32,888 & 1,827 & 1,828 & LP & 2020 \\
    CoDEx-M \cite{codex} & 11,941 & 50    & 185,584 & 10,310 & 10,311 & LP & 2020 \\
    CoDEx-L \cite{codex} & 45,869 & 69    & 551,193 & 30,622 & 30,622 & LP & 2020 \\
    Wikidata5M-transductive \cite{kepler} & 4,594,485 & 822   & 20,614,279 & 5,163 & 5,133 & LP & 2021 \\
    Wikidata5M-inductive \cite{kepler} & 4,594,458 & 822   & 20,496,514 & 6,699 & 6,894 & LP(ZS) & 2021 \\
    Diabetes \cite{diabetes} & 7,886 & 67    & 56,830 & 1,344 & 1,936 & LP    & 2022 \\
    \hline
    \end{tabularx}%
  }
  \caption{Statistics on Datasets.}
  \label{tab:datasets}%
\end{table}%

\subsection{Datasets}
To facilitate a systematic and comparative analysis of existing KRL methods, this section summarizes the datasets commonly used in previous studies, which serve as a foundation for evaluating the generalizability and effectiveness of different methods. The key statistics for each dataset are detailed in Table \ref{tab:datasets}. This overview aims to consolidate the dispersed information and provide a unified reference for subsequent analysis.

\subsection{Metrics}
This section employs the commonly understood definitions of accuracy, precision, recall, and F1-score, which are widely recognized metrics for evaluating classification tasks. In addition to these, we introduce metrics specifically designed for ranking tasks: Mean Rank (MR), Mean Reciprocal Rank (MRR), and Hits@K.

\begin{itemize}
    \item \textbf{Mean Rank} is a metric for ranking tasks, such as knowledge graph completion. It represents the average rank position of the correct entity or relation in the predicted list. Lower MR values indicate better performance:
    \begin{equation}
        \text{MR} = \frac{1}{N} \sum_{i=1}^{N} \text{rank}_i
    \end{equation}
    \item \textbf{Mean Reciprocal Rank} evaluates the effectiveness of a ranking algorithm. It is the average of the reciprocal ranks of the correct answers, useful for information retrieval and question answering tasks. Higher MRR values indicate better performance:
    \begin{equation}
        \text{MRR} = \frac{1}{N} \sum_{i=1}^{N} \frac{1}{\text{rank}_i}
    \end{equation}
    \item \textbf{Hits@K} measures the proportion of correct answers appearing in the top $K$ predicted results, commonly used in ranking tasks and knowledge graph completion. Higher Hits@K values indicate better performance:
    \begin{equation}
        \text{Hits@K} = \frac{1}{N} \sum_{i=1}^{N} \mathbb{1}(\text{rank}_i \leq K)
    \end{equation}
\end{itemize}

\subsection{Downstream Tasks}
\begin{figure*}
\centering
\includegraphics[width=\textwidth]{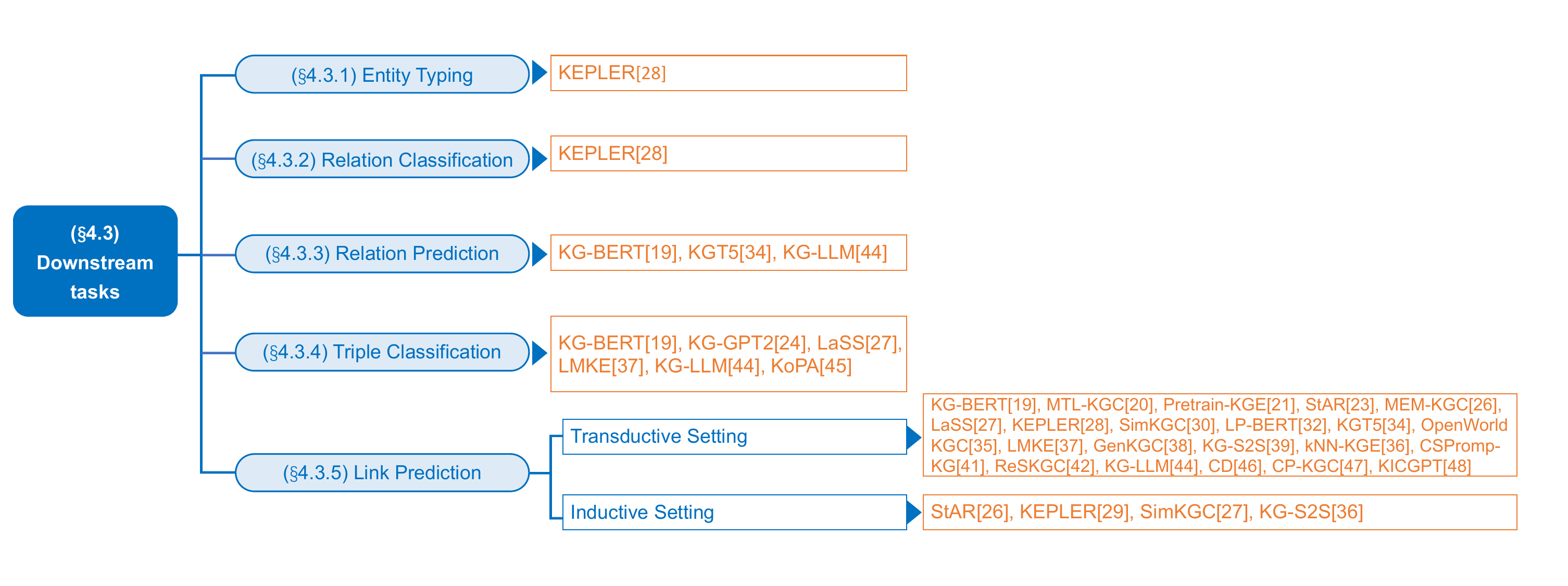}
\caption{An overview of downstream tasks achievable by each LLM-enhanced KRL model.}
\label{fig.experiment}
\end{figure*}

Building on the dataset overview, this section focuses on downstream tasks commonly used to evaluate KRL models. By summarizing these tasks, we propose a comparative framework that can illuminate the strengths and weaknesses of different methods under various conditions. While we systematically organize findings from prior studies (Fig. \ref{fig.experiment}), it is worth noting that certain metrics or settings may not always align, making direct comparisons of some models less straightforward. Nevertheless, we hope these insights can foster a deeper understanding of current challenges in the field.

\subsubsection{Entity Type Classification}
\textit{Definition 1:} \textbf{Entity Typing (ET).} ET is a fundamental task that involves assigning predefined types or categories to entities mentioned within a text. Formally, given a text corpus $T$ and a set of entities $E$, the goal is to determine a mapping $f: E \rightarrow C$, where $C$ is the set of predefined types. Specifically, for an entity $e \in E$ appearing in a context $t \in T$, the task is to predict its type $c \in C$. This can be expressed as:

\begin{equation}
f(e, t) = c, \quad \text{where } e \in E, \ t \in T, \ \text{and } c \in C.
\end{equation}

\begin{table}[htbp]
  \centering
  \scalebox{0.88}{
    \begin{tabular}{c|cc|ccc|ccc}
    \toprule
    \multirow{2}[1]{*}{Model} & \multirow{2}[1]{*}{LLM-Enhanced} & \multirow{2}[1]{*}{Year} & \multicolumn{3}{c|}{FIGER} & \multicolumn{3}{c}{Open Entity} \\
          &       &       & Acc   & Macro & Micro & P     & R     & F1 \\
    \midrule
    NFGEC \cite{nfgec} & × & 2016  & \underline{55.60}  & 75.15 & \underline{71.73} & 68.80  & 53.30  & 60.10 \\
    UFET \cite{open_entity}  & × & 2018  & -     & -     & -     & 77.40  & 60.60  & 68.00 \\
    BERT \cite{bert} & × & 2018 & 52.04 & \underline{75.16} & 71.63 & 76.37 & 70.96 & 73.56 \\
    RoBERTa \cite{roberta} & × & 2019  & -     & -     & -     & 77.40  & \underline{73.60}  & 75.40 \\
    ERNIE \cite{ernie} & × & 2019  & \textbf{57.19} & \textbf{76.51} & \textbf{73.39} & \underline{78.42} & 72.90  & 75.56 \\
    KnowBert \cite{knowbert} & × & 2019  & -     & -     & -     & \textbf{78.70}  & 72.70  & \underline{75.60} \\
    KEPLER \cite{kepler} & \checkmark & 2021  & -     & -     & -     & 77.80  & \textbf{74.60}  & \textbf{76.20} \\
    \bottomrule
    \end{tabular}%
  }
  \caption{Experimental statistics under the ET subtask.}
  \label{tab:et_performance}%
\end{table}%

Table \ref{tab:et_performance} presents experimental results for ET on two widely used datasets, FIGER and Open Entity. Several notable trends emerge:

Traditional feature-based methods such as NFGEC and UFET demonstrate early successes in capturing essential entity information, yet their performance lags behind the more recent transformer-based approaches. Models built upon LLM, including BERT and RoBERTa, generally achieve higher accuracy and F1 scores, showcasing the importance of contextual embeddings derived from extensive text corpora.

Methods that incorporate external knowledge sources (e.g., ERNIE, KnowBert, and KEPLER) further improve ET performance by injecting entity-related information into their representations. For instance, ERNIE achieves impressive results on FIGER (e.g., reaching 57.19\% accuracy and 73.39\% in micro-F1), highlighting the benefits of entity-level knowledge. Similarly, KEPLER shows a slight edge in F1 score (76.20\%) over comparable models on Open Entity, suggesting that combining textual embeddings with knowledge graph features can enhance model robustness.

Not all models report a full set of comparable metrics. For example, KEPLER does not provide FIGER results. Meanwhile, KnowBert on Open Entity lacks certain detailed scores. Such reporting gaps limit direct comparisons across all methods. Hence, while we observe a consistent performance boost from integrating external knowledge into LLMs, we cannot definitively conclude that knowledge-augmented models universally dominate all ET benchmarks.

These findings underscore both the effectiveness of LLMs in capturing rich context and the potential gains from fusing textual and structural information. To facilitate more comprehensive evaluations, future research may focus on establishing consistent experimental protocols and shared benchmarks, enabling a deeper understanding of how ET models perform across diverse data domains and label distributions.

\subsubsection{Relation Classification}

\textit{Definition 2:} \textbf{Relation Classification (RC).} RC is a crucial task that involves identifying the semantic relationship between a pair of entities within a sentence. Formally, given a sentence $S$ containing two entities $e_1$ and $e_2$, and a predefined set of relation types $R$, the task is to determine a mapping $f: (e_1, e_2, S) \rightarrow r$, where $r \in R$ is the relation type. Specifically, for a pair of entities $(e_1, e_2)$ in the context of sentence $S$, the goal is to predict the relation $r$. This can be expressed as:

\begin{equation}
f(e_1, e_2, S) = r, \quad \text{where } e_1, e_2 \in E, \ S \in T, \ \text{and } r \in R.
\end{equation}

\begin{table}[htbp]
  \centering
  \scalebox{0.88}{
    \begin{tabular}{c|cc|ccc|ccc}
    \toprule
    \multirow{2}[1]{*}{Model} & \multirow{2}[1]{*}{LLM-Enhanced} & \multirow{2}[1]{*}{Year} & \multicolumn{3}{c|}{FewRel} & \multicolumn{3}{c}{TACRED} \\
          &       &       & P & R & F1    & P & R & F1 \\
    \midrule
    CNN \cite{cnn}   & × & 2015  & 69.51 & 69.64 & 69.35 & 70.30  & 54.20  & 61.20 \\
    PA-LSTM \cite{pa-lstm} & × & 2017  & -     & -     & -     & 65.70  & 64.50  & 65.10 \\
    C-GCN \cite{c-gcn} & × & 2018  & -     & -     & -     & 69.90  & 63.30  & 66.40 \\
    BERT\_Base \cite{bert} & × & 2018  & \underline{85.05} & \underline{85.11} & \underline{84.89} & 67.23 & 64.81 & 66.00 \\
    BERT\_Large \cite{bert} & × & 2018  & -     & -     & -     & -     & -     & 70.10 \\
    RoBERTa \cite{roberta} & × & 2019  & -     & -     & -     & 70.40  & 71.10  & 70.70 \\
    MTB \cite{mtb}   & × & 2019  & -     & -     & -     & -     & -     & \underline{71.50} \\
    ERNIE \cite{ernie} & × & 2019  & 88.49 & \textbf{88.44} & \textbf{88.32} & 69.97 & 66.08 & 67.97 \\
    KnowBert \cite{knowbert} & × & 2019  & -     & -     & -     & \textbf{71.60}  & \underline{71.40}  & \underline{71.50} \\
    KEPLER \cite{kepler} & \checkmark & 2021  & -     & -     & -     & \underline{71.50}  & \textbf{72.50}  & \textbf{72.00} \\
    SemGL \cite{semgl} & × & 2024  &  \textbf{95.11}     & -     & -     & -  & -  & - \\
    \bottomrule
    \end{tabular}%
  }
  \caption{Experimental statistics under the RC subtask.}
  \label{tab:rc_performance}%
\end{table}%

Table \ref{tab:rc_performance} summarizes the performance of various RC models on the FewRel and TACRED datasets.

Early methods such as CNN\,\cite{cnn} rely on convolutional feature extractors and achieve an F1 score of 69.35\%, illustrating the utility of neural architectures in capturing local text patterns. When large-scale pre-training is introduced, transformer-based models markedly improve results. BERT\_Base\,\cite{bert} raises F1 to 84.89\%, while ERNIE\,\cite{ernie}, which incorporates entity-level knowledge, attains 88.32\%. These gains highlight the value of infusing external knowledge for more nuanced relation modeling. SemGL\,\cite{semgl} further reports high precision (95.11\%) on FewRel, but lacks complete recall and F1 metrics, preventing direct, comprehensive comparisons.

On TACRED, CNN and PA-LSTM exhibit moderate performance, trailing behind graph-based encoders like C-GCN\,\cite{c-gcn}, which benefits from structured dependency information (F1\,=\,66.40\%). Transformer-based systems perform better still: BERT\_Large hits 70.10\% F1, and RoBERTa\,\cite{roberta} and MTB\,\cite{mtb} exceed 70.70\% F1. Subsequent models leverage external knowledge to enhance representations. For example, KnowBert\,\cite{knowbert} and ERNIE incorporate knowledge bases to surpass 71\% F1. The current best results among the listed systems come from KEPLER\,\cite{kepler}, an LLM-enhanced approach that achieves an F1 of 72.00\%, illustrating the added benefit of entity-description alignment in capturing relational patterns.

Overall, transformer-based encoders dominate, particularly when combined with external knowledge. Traditional CNN or RNN architectures, despite strong early results, struggle to keep pace with LLMs. Meanwhile, knowledge-aware methods such as ERNIE, KnowBert, and KEPLER showcase further advances by seamlessly blending textual and structural cues. However, direct model comparisons are partially hampered by incomplete or inconsistent reporting across datasets.

\subsubsection{Relation Prediction}

\textit{Definition 3:} \textbf{Relation Prediction (RP).} RP is a task in KG completion that aims to predict the relationship between two entities within a knowledge graph. Given a set of entities $E$ and a set of possible relations $R$, the objective is to predict the relation $r \in R$ between a given pair of head entity $h \in E$ and tail entity $t \in E$. Formally, the task can be defined as:

\begin{equation}
f: E \times E \rightarrow R
\end{equation}

For a given pair of entities $(h, t)$, the task is to determine the most likely relation $r$ that connects them:

\begin{equation}
r = f(h, t), \quad \text{where } h, t \in E \text{ and } r \in R.
\end{equation}

\begin{table}[htbp]
  \centering
  \scalebox{0.88}{
    \begin{tabular}{c|cc|cc|c}
    \toprule
    \multirow{2}[1]{*}{Model} & \multirow{2}[1]{*}{LLM-Enhanced} & \multirow{2}[1]{*}{Year} & \multicolumn{2}{c}{FB15K} & YAGO3-10-100 \\
          &       &       & MR    & Hits@1 & Hits@1 \\
    \midrule
    TransE \cite{fb15k} & × & 2013  & 2.5   & 84.3  & - \\
    TransR \cite{transr} & × & 2015  & 2.1   & 91.6  & - \\
    DKRL \cite{dkrl}  & × & 2016  & 2.0     & 90.8  & - \\
    TKRL \cite{tkrl}  & × & 2016  & \underline{1.7}   & 92.8  & - \\
    PTransE \cite{ptranse} & × & 2015  & \textbf{1.2}   & 93.6  & - \\
    SSP \cite{ssp}   & × & 2017  & \textbf{1.2}   & -     & - \\
    ProjE \cite{proje} & × & 2017  & \textbf{1.2}   & \underline{95.7}  & - \\
    KG-BERT \cite{kg-bert} & \checkmark & 2019  & \textbf{1.2}   & \textbf{96.0}    & - \\
    KGT5 \cite{kgt5}  & \checkmark & 2022  & -     & -     & \underline{60} \\
    ChatGPT & × & 2023  & -     & -     & 39 \\
    GPT-4 & × & 2023  & -     & -     & 56 \\
    LLaMA-7B \cite{llama2} & × & 2023  & -     & -     & 13 \\
    KG-LLM \cite{kg-llm} & \checkmark & 2023  & -     & -     & \textbf{71} \\
    \bottomrule
    \end{tabular}%
  }
  \caption{Experimental statistics under the RP subtask.}
  \label{tab:rp_performance}%
\end{table}%

Table \ref{tab:rp_performance} reports representative results on two benchmark datasets, FB15K and YAGO3-10-100.

Classical translational methods, such as TransE and PTransE, already achieve high performance on FB15K (MR of around 1.2 and Hits@1 between 93.6\% and 95.7\%). Among these, ProjE attains Hits@1 of 95.7\%, and KG-BERT further improves to 96.0\%. This suggests that incorporating textual information can help refine relational reasoning on FB15K. However, it remains unclear how these same translational baselines would perform on more complex or diverse benchmarks, as most of them have not been evaluated on YAGO3-10-100.

Compared to FB15K, YAGO3-10-100 presents richer semantics and a more varied entity space. Older translational methods provide no results on this dataset, making cross-method comparisons incomplete. Nonetheless, several LLMs and fine-tuned approaches have been evaluated. KGT5, an encoder-decoder model adapted for KGC, reports Hits@1 of 60\%. General-purpose LLMs (ChatGPT, GPT-4, LLaMA-7B) yield Hits@1 scores of 39\%, 56\%, and 13\%, respectively. Though they demonstrate some ability to handle relation inference, their out-of-the-box performance lags behind specialized KGC models. KG-LLM, a decoder-based model instruction-tuned for KGC, achieves the highest known Hits@1 of 71\%. This underscores that tailoring large models to KGC tasks—through specialized prompts, fine-tuning objectives, or both—can unlock better relational predictions.

Because earlier structural embeddings (e.g., TransE, DistMult) have not been tested on YAGO3-10-100, it is difficult to determine whether their strong performance on FB15K would generalize to YAGO3-10-100’s more complex patterns. Similarly, while KGT5 performs reasonably well, the absence of matching baselines (in terms of model type or dataset coverage) prevents a direct head-to-head comparison.

Overall, on FB15K, incorporating textual context (e.g., KG-BERT) can outperform purely structural methods, suggesting that text-informed embeddings bolster relation prediction. On YAGO3-10-100, the top performance belongs to KG-LLM, closely followed by KGT5, whereas general-purpose LLMs without specialized tuning show lower Hits@1. However, these results remain challenging to fully interpret due to the lack of consistent evaluations across older translation-based models and different LLMs on YAGO3-10-100. A more uniform experimental setup across both classical and LLM-enhanced methods on larger, semantically diverse datasets would help clarify the extent to which textual integration, instruction-tuning, or architectural choices most significantly drive relation prediction improvements.

\subsubsection{Triple Classification}

\textit{Definition 4:} \textbf{Triple Classification (TC).} TC is a knowledge graph completion task, aimed at determining the plausibility of a given knowledge graph triple. Formally, given a set of entities $E$, a set of relations $R$, and a knowledge graph $K \subseteq E \times R \times E$ consisting of triples $(h, r, t)$, where $h$ is the head entity, $r$ is the relation, and $t$ is the tail entity, the task is to classify each triple as either true or false. The classification function can be defined as:

\begin{equation}
f: E \times R \times E \rightarrow \{0, 1\}
\end{equation}

where $f(h, r, t) = 1$ indicates that the triple $(h, r, t)$ is plausible (true), and $f(h, r, t) = 0$ indicates it is implausible (false).

\begin{table}[htbp]
  \centering
  \scalebox{0.88}{
    \begin{tabular}{c|cc|c|c|c|c|c}
    \toprule
    \multirow{2}[1]{*}{Model} & \multirow{2}[1]{*}{LLM-Enhanced} & \multirow{2}[1]{*}{Year} & WN11  & FB13  & UMLS  & CoDEX-S & FB15k-237N \\
          &       &       & Acc & Acc & Acc & Acc & Acc \\
    \midrule
    NTN \cite{wn11_fb13}   & × & 2013  & 86.2  & 90.0    & -     & -     & - \\
    TransE \cite{fb15k} & × & 2013  & 75.9  & 81.5  & 78.1  & 72.1  & - \\
    DistMult \cite{distmult} & × & 2014  & 87.1  & 86.2  & 86.8  & 66.8  & - \\
    ComplEx \cite{complex} & × & 2016  & -     & -     & 90.8  & 67.8  & 65.7 \\
    RotatE \cite{rotate} & × & 2019  & -     & -     & 92.1  & 75.7  & 68.5 \\
    KG-BERT \cite{kg-bert} & \checkmark & 2019  & 93.5  & 90.4  & 89.7  & 77.3  & 56.0 \\
    KG-GPT2 \cite{kg-gpt2} & \checkmark & 2021  & 85.0    & 89.0    & -     & -     & - \\
    LaSS \cite{lass}  & \checkmark & 2021  & \underline{94.5}  & \textbf{91.8}  & -     & -     & - \\
    KGT5 \cite{kgt5}  & \checkmark & 2022  & 72.8  & 66.3  & -     & -     & - \\
    LMKE \cite{lmke}  & \checkmark & 2022  & -     & \underline{91.7}  & \underline{92.4}  & -     & - \\
    GPT-3.5 & × & 2023  & -     & -     & 67.6  & 54.7  & 60.2 \\
    LLaMA-7B \cite{llama2} & × & 2023  & 21.1  & 9.1   & -     & -     & - \\
    LLaMA-13B \cite{llama2} & × & 2023  & 28.1  & 17.6  & -     & -     & - \\
    KG-LLM \cite{kg-llm} & \checkmark & 2023  & \textbf{95.6}  & 90.2  & 86.0    & \underline{80.3}  & \textbf{80.5} \\
    KoPA \cite{kopa}  & \checkmark & 2023  & -     & -     & \textbf{92.6}  & \textbf{82.7}  & \underline{77.7} \\
    \bottomrule
    \end{tabular}%
  }
  \caption{Experimental statistics under the TC subtask.}
  \label{tab:tc_performance}%
\end{table}%

Table \ref{tab:tc_performance} summarizes the reported accuracy of representative models on five widely used datasets of TC task: WN11, FB13, UMLS, CoDEx-S, and FB15k-237N.

One striking observation is that LLM-enhanced methods generally achieve higher accuracy. KG-LLM attains the highest reported accuracy on WN11 (95.6\%), while KoPA achieves 82.7\% on CoDEx-S; both outperform earlier purely structural methods. This boost in performance is closely tied to the ability of LLM-based approaches to exploit textual semantics in addition to graph structure, thereby capturing subtle relational nuances. Despite this overall trend favoring LLM-based approaches, certain structural models still perform competitively on specific datasets. RotatE, for example, reaches 92.1\% accuracy on UMLS, suggesting that, for simpler or smaller-scale data, well-tuned classical methods may remain strong contenders.

Nevertheless, the impact of dataset characteristics on model performance is also evident. WN11 and FB13, which feature moderate scales and relation complexity, see large gains for encoder-based methods such as KG-BERT and LMKE, each exceeding 90\% accuracy. By contrast, FB15k-237N and CoDEx-S often include more diverse or fine-grained relations, where decoder-based architectures like KG-LLM and KoPA can excel. Notably, KG-LLM reaches 80.5\% accuracy on FB15k-237N, surpassing older techniques that lack the benefits of large-scale pretrained linguistic knowledge. Still, generic large language models (e.g., GPT-3.5, LLaMA) may struggle without knowledge graph alignment: they show markedly lower accuracy (sometimes below 30\% on WN11), indicating that additional fine-tuning or structural alignment (e.g., prefix adapters, prompt engineering) is crucial.

The comparisons remain incomplete due to missing results in the literature. Certain classic embedding models (e.g., TransE, ComplEx) have not been evaluated on newer datasets such as CoDEx-S or specific variants like FB15k-237N, while some LLM-based methods do not report outcomes on FB13 or UMLS. This lack of uniform baselines means that one cannot universally conclude that LLM-enhanced models will always outperform structural methods across all scenarios. Where data does exist, however, trends lean strongly in favor of the methods integrating pretrained language models, especially for more complex or semantically rich datasets.

While incorporating large-scale textual semantics confers clear benefits on triple classification, practical deployment of LLM-based methods still demands considerable computational overhead. Moreover, the promising results observed thus far do not fully generalize to every domain or dataset in the absence of more exhaustive benchmarking. Consequently, model choice should be guided by task-specific considerations such as dataset size, relation complexity, resource constraints, and the availability of textual descriptions. Although LLM-driven approaches show strong potential, systematic evaluations under varied conditions are essential to confirm these methods’ broader applicability and to assess how best to reconcile performance gains with feasibility in real-world scenarios.

\subsubsection{Link Prediction}
\textit{Definition 5:} \textbf{Link Prediction (LP).} LP is a core task in KG research, aiming to infer missing links between entities. Formally, given a knowledge graph $G = (E, R, T)$ where $E$ represents the set of entities, $R$ represents the set of relations, and $T \subseteq E \times R \times E$ represents the set of triples (links), the task is to predict the plausibility of a new triple $(h, r, t)$ being true, where $h, t \in E$ and $r \in R$. The goal is to find a scoring function $f: E \times R \times E \rightarrow \mathbb{R}$ that assigns a high score to plausible triples and a low score to implausible ones. This can be expressed as:

\begin{equation}
f(h, r, t) \approx \text{True}, \quad \text{for } (h, r, t) \in G.
\end{equation}

\begin{table}[!htbp]
  \centering
  \scalebox{0.88}{
    \begin{tabular}{c|cc|ccccc}
    \toprule
    \multirow{2}[1]{*}{Model} & \multirow{2}[1]{*}{LLM-Enhanced} & \multirow{2}[1]{*}{Year} & \multicolumn{5}{c}{WN18RR} \\
          &       &       & MR    & MRR   & Hits@1 & Hits@3 & Hits@10 \\
    \midrule
    TransE \cite{fb15k} & × & 2013  & 2300  & 24.3  & 4.3   & 44.1  & 53.2 \\
    TransH \cite{transh} & × & 2014  & 2524  & -     & -     & -     & 50.3 \\
    DistMult \cite{distmult} & × & 2014  & 3704  & 44.4  & 41.2  & 47.0    & 50.4 \\
    TransR \cite{transr} & × & 2015  & 3166  & -     & -     & -     & 50.7 \\
    TransD \cite{transd} & × & 2015  & 2768  & -     & -     & -     & 50.7 \\
    ComplEx \cite{complex} & × & 2016  & 3921  & 44.9  & 40.9  & 46.9  & 53.0 \\
    ConvE \cite{wn18rr_umls_yago3-10} & × & 2018  & 4464  & 45.6  & 41.9  & 47.0    & 53.1 \\
    ConvKB \cite{convkb} & × & 2018  & 2554  & 24.9  & -     & -     & 52.5 \\
    R-GCN \cite{r-gcn} & × & 2018  & 6700  & 12.3  & 8.0     & 13.7  & 20.7 \\
    KBGAN \cite{kbgan} & × & 2018  & -     & 21.5  & -     & -     & 48.1 \\
    RotatE \cite{rotate} & × & 2019  & 3340  & 47.6  & 42.8  & 49.2  & 57.1 \\
    KBAT \cite{kbat}  & × & 2019  & 1921  & 41.2  & -     & -     & 55.4 \\
    CapsE \cite{capse} & × & 2019  & 718   & 41.5  & -     & -     & 55.9 \\
    QuatE \cite{quate} & × & 2019  & 3472  & 48.1  & 43.6  & 50.0    & 56.4 \\
    TuckER \cite{tucker} & × & 2019  & -     & 47.0    & 44.3  & 48.2  & 52.6 \\
    HAKE \cite{hake}  & × & 2019  & -     & 49,7  & 45.2  & 51.6  & 58.2 \\
    AttH \cite{atth}  & × & 2020  & -     & 48.6  & 44.3  & 49.9  & 57.3 \\
    REFE \cite{refe}  & × & 2020  & -     & -     & -     & -     & 56.1 \\
    GAATs \cite{gaats} & × & 2020  & 1270  & -     & -     & -     & 60.4 \\
    Complex-DURA \cite{complex-dura} & × & 2020  & 57  & -     & -     & -     & - \\
    DensE \cite{dense} & × & 2020  & 3052  & 49.1  & 44.3  & 50.8  & 57.9 \\
    LineaRE \cite{lineare} & × & 2020  & 1644  & 49.5  & 45.3  & 50.9  & 57.8 \\
    RESCAL-DURA \cite{rescal-dura} & × & 2020  & -     & 49.8  & 45.5  & -     & 57.7 \\
    CompGCN \cite{compgcn} & × & 2020  & -     & 47.9  & 44.3  & 49.4  & 54.6 \\
    NePTuNe \cite{neptune} & × & 2021  & -     & -     & -     & -     & 55.7   \\
    ComplEx-N3-RP \cite{complex-n3-rp} & × & 2021  & -     & -     & -     & -     & 58.0 \\
    ConE \cite{cone}  & × & 2021  & -     & 49.6  & 45.3  & 51.5  & 57.9 \\
    Rot-Pro \cite{rot-pro} & × & 2021  & -     & 45.7  & 39.7  & 48.2  & 57.7 \\
    QuatDE \cite{quatde} & × & 2021  & 1977  & 48.9  & 43.8  & 50.9  & 58.6 \\
    NBFNet \cite{nbfnet} & × & 2021  & -     & 55.1  & 49.7  & -     & 66.6 \\
    \midrule
    KG-BERT \cite{kg-bert} & \checkmark & 2019  & 97    & 21.6  & 4.1   & 30.2  & 52.4 \\
    MTL-KGC \cite{mtl-kgc} & \checkmark & 2020  & 89    & 33.1  & 20.3  & 38.3  & 59.7 \\
    Pretrain-KGE \cite{pretrain-kge} & \checkmark & 2020  & -     & 48.8  & 43.7  & 50.9  & 58.6 \\
    StAR \cite{star}  & \checkmark & 2021  & \underline{51}    & 40.1  & 24.3  & 49.1  & 70.9 \\
    MEM-KGC \cite{mem-kgc} & \checkmark & 2021  & -     & 57.2  & 48.9  & 62.0    & 72.3 \\
    LaSS \cite{lass}  & \checkmark & 2021  & \textbf{35}    & -     & -     & -     & 78.6 \\
    SimKGC \cite{simkgc} & \checkmark & 2021  & -     & \underline{66.7}  & \underline{58.8}  & \textbf{72.1}  & \textbf{80.5} \\
    LP-BERT \cite{lp-bert} & \checkmark & 2022  & 92    & 48.2  & 34.3  & 56.3  & 75.2 \\
    KGT5 \cite{kgt5}  & \checkmark & 2022  & -     & 54.2  & 50.7  & -     & 60.7 \\
    OpenWorld KGC \cite{openworld-kgc} & \checkmark & 2022  & -     & 55.7  & 47.5  & 60.4  & 70.4 \\
    LMKE \cite{lmke}  & \checkmark & 2022  & 79    & 61.9  & 52.3  & \underline{67.1}  & 78.9 \\
    GenKGC \cite{genkgc} & \checkmark & 2022  & -     & -     & 28.7  & 40.3  & 53.5 \\
    KG-S2S \cite{kgs2s} & \checkmark & 2022  & -     & 57.4  & 53.1  & 59.5  & 66.1 \\
    kNN-KGE \cite{knn-kge} & \checkmark & 2023  & -     & 57.9  & 52.5  & -     & - \\
    CSPromp-KG \cite{csprom-kg} & \checkmark & 2023  & -     & 57.5  & 52.2  & 59.6  & 67.8 \\
    GPT-3.5 & × & 2023  & -     & -     & 19.0    & -     & - \\
    LLaMA-7B \cite{llama2} & × & 2023  & -     & -     & 8.5   & -     & - \\
    LLaMA-13B \cite{llama2} & × & 2023  & -     & -     & 10.0    & -     & - \\
    KG-LLM \cite{kg-llm} & \checkmark & 2023  & -     & -     & 25.6 & -     & - \\
    CD \cite{cd}    & \checkmark & 2024  & -     & 57.6  & 52.6  & 60.7  & 67.2 \\
    CP-KGC \cite{cp-kgc} & \checkmark & 2024  & -     & \textbf{67.3}  & \textbf{59.9}  & \textbf{72.1}  & \underline{80.4} \\
    KICGPT \cite{kicgpt} & \checkmark & 2024  & -     & 56.4  & 47.8  & 61.2  & 67.7 \\
    \bottomrule
    \end{tabular}%
  }
  \caption{Experimental statistics of WN18RR dataset under the LP subtask in transductive settings.}
  \label{tab:lp_wn18rr}%
\end{table}%

\begin{table}[!htbp]
  \centering
  \scalebox{0.88}{
    \begin{tabular}{c|cc|ccccc}
    \toprule
    \multirow{2}[1]{*}{Model} & \multirow{2}[1]{*}{LLM-Enhanced} & \multirow{2}[1]{*}{Year} & \multicolumn{5}{c}{FB15k-237} \\
          &       &       & MR    & MRR   & Hits@1 & Hits@3 & Hits@10 \\
    \midrule
    TransE \cite{fb15k} & × & 2013  & 223   & 27.9  & 19.8  & 37.6  & 47.4 \\
    TransH \cite{transh} & × & 2014  & 255   & -     & -     & -     & 48.6 \\
    DistMult \cite{distmult} & × & 2014  & 411   & 28.1  & 19.9   & 30.1  & 44.6 \\
    TransR \cite{transr} & × & 2015  & 237   & -     & -     & -     & 51.1 \\
    TransD \cite{transd} & × & 2015  & 246   & -     & -     & -     & 48.4 \\
    ComplEx \cite{complex} & × & 2016  & 508   & 27.8  & 19.4  & 29.7  & 45.0 \\
    ConvE \cite{wn18rr_umls_yago3-10} & × & 2018  & 245   & 31.2  & 22.5  & 34.1  & 49.7 \\
    ConvKB \cite{convkb} & × & 2018  & 257   & 24.3  & -     & -     & 51.7 \\
    R-GCN \cite{r-gcn} & × & 2018  & 600   & 16,4  & 10.0    & 18.1  & 41.7 \\
    KBGAN \cite{kbgan} & × & 2018  & -     & 27.7  & -     & -     & 45.8 \\
    RotatE \cite{rotate} & × & 2019  & 177   & 33.8  & 24.1  & 37.5  & 53.3 \\
    KBAT \cite{kbat}  & × & 2019  & 270   & 15.7  & -     & -     & 33.1 \\
    CapsE \cite{capse} & × & 2019  & 403   & 15.0    & -     & -     & 35.6 \\
    QuatE \cite{quate} & × & 2019  & 176   & 31.1  & 22.1  & 34.2  & 49.5 \\
    TuckER \cite{tucker} & × & 2019  & -     & 35.8  & 26.6  & 39.4  & 54.4 \\
    HAKE \cite{hake}  & × & 2019  & -     & 34.6  & 25.0    & 38.1  & 54.2 \\
    AttH \cite{atth}  & × & 2020  & -     & 34.8  & 25.2  & 38.4  & 54.0 \\
    REFE \cite{refe}  & × & 2020  & -     & -     & -     & -     & 54.1 \\
    GAATs \cite{gaats} & × & 2020  & 187   & -     & -     & -     & 65.0 \\
    Complex-DURA \cite{complex-dura} & × & 2020  & \textbf{56}    & -     & -     & -     & - \\
    DensE \cite{dense} & × & 2020  & 169   & 34.9  & 25.6  & 38.4  & 53.5 \\
    LineaRE \cite{lineare} & × & 2020  & 155   & 35.7  & 26.4  & 39.1  & 54.5 \\
    RESCAL-DURA \cite{rescal-dura} & × & 2020  & -     & 36.8  & 27.6  & -     & 55.0 \\
    CompGCN \cite{compgcn} & × & 2020  & -     & 35.5  & 26.4  & 39.0    & 53.5 \\
    NePTuNe \cite{neptune} & × & 2021  & -     & -     & -     & -     & 54.7 \\
    ComplEx-N3-RP \cite{complex-n3-rp} & × & 2021  & -     & -     & -     & -     & 56.8 \\
    ConE \cite{cone}  & × & 2021  & -     & 34.5  & 24.7  & 38.1  & 54.0 \\
    Rot-Pro \cite{rot-pro} & × & 2021  & -     & 34.4  & 24.6  & 38.3  & 54.0 \\
    QuatDE \cite{quatde} & × & 2021  & \underline{90}    & 36.5  & 26.8  & \underline{40.0}    & \underline{56.3} \\
    NBFNet \cite{nbfnet} & × & 2021  & -     & \textbf{41.5}  & 32.1  & -     & \textbf{59.9} \\
    \midrule
    KG-BERT \cite{kg-bert} & \checkmark & 2019  & 153   & 23.7  & 16.9  & 26.0    & 42.7 \\
    MTL-KGC \cite{mtl-kgc} & \checkmark & 2020  & 132   & 26.7  & 17.2  & 29.8  & 45.8 \\
    Pretrain-KGE \cite{pretrain-kge} & \checkmark & 2020  & -     & 35.0    & 25.0    & 38.4  & 55.4 \\
    StAR \cite{star}  & \checkmark & 2021  & 1117  & 29.6  & 20.5  & 32.2  & 48.2 \\
    MEM-KGC \cite{mem-kgc} & \checkmark & 2021  & -     & 34.9  & 26.0    & 38.2  & 52.4 \\
    LaSS \cite{lass}  & \checkmark & 2021  & 108   & -     & -     & -     & 53.3 \\
    SimKGC \cite{simkgc} & \checkmark & 2021  & -     & 33.6  & 24.9  & 36.2  & 51.1 \\
    LP-BERT \cite{lp-bert} & \checkmark & 2022  & 154   & 31.0    & 22.3  & 33.6  & 49.0 \\
    KGT5 \cite{kgt5}  & \checkmark & 2022  & -     & 34.3  & 25.2  & -     & 37.7 \\
    OpenWorld KGC \cite{openworld-kgc} & \checkmark & 2022  & -     & 34.6  & 25.3  & 38.1  & 53.1 \\
    LMKE \cite{lmke}  & \checkmark & 2022  & 141   & 30.6  & 21.8  & 33.1  & 48.4 \\
    GenKGC \cite{genkgc} & \checkmark & 2022  & -     & -     & 19.2  & 35.5  & 43.9 \\
    KG-S2S \cite{kgs2s} & \checkmark & 2022  & -     & 33.6  & 25.7  & 37.3  & 49.8 \\
    kNN-KGE \cite{knn-kge} & \checkmark & 2023  & -     & 28.0    & \textbf{37.3}  & -     & - \\
    CSPromp-KG \cite{csprom-kg} & \checkmark & 2023  & -     & 35.8  & 26.9  & 39.3  & 53.8 \\
    GPT-3.5 & × & 2023  & -     & -     & 23.7  & -     & - \\
    CP-KGC \cite{cp-kgc} & \checkmark & 2024  & -     & 33.8  & 25.1  & 36.5  & 51.6 \\
    KICGPT \cite{kicgpt} & \checkmark & 2024  & -     & \underline{41.2}  & \underline{32.7}  & \textbf{44.8}  & 55.4 \\
    \bottomrule
    \end{tabular}%
  }
  \caption{Experimental statistics of FB15k-237 dataset under the LP subtask in transductive settings.}
  \label{tab:lp_fb15k-237}%
\end{table}%

\begin{table}[!htbp]
  \centering
  \scalebox{0.88}{
    \begin{tabular}{c|cc|cc|cccc}
    \toprule
    \multirow{2}[1]{*}{Model} & \multirow{2}[1]{*}{LLM-Enhanced} & \multirow{2}[1]{*}{Year} & \multicolumn{2}{c|}{UMLS} & \multicolumn{4}{c}{Wikidata5M} \\
          &       &       & MR    & Hits@10 & MRR   & Hits@1 & Hits@3 & Hits@10 \\
    \midrule
    TransE \cite{fb15k} & × & 2013  & 1.84  & 98.9  & 25.3  & 17.0    & 31.1  & 39.2 \\
    TransH \cite{transh} & × & 2014  & 1.80   & 99.5  & -     & -     & -     & - \\
    DistMult \cite{distmult} & × & 2014  & 5.52  & 84.6  & 25.3  & 20.9  & 27.8  & 33.4 \\
    TransR \cite{transr} & × & 2015  & 1.81  & 99.4  & -     & -     & -     & - \\
    TransD \cite{transd} & × & 2015  & 1.71  & 99.3  & -     & -     & -     & - \\
    ComplEx \cite{complex} & × & 2016  & 2.59  & 96.7  & 30.8  & 25.5  & -     & 39.8 \\
    DKRL \cite{dkrl}  & × & 2016  & -     & -     & 16.0    & 12.0    & 18.1  & 22.9 \\
    ConvE \cite{wn18rr_umls_yago3-10} & × & 2018  & 1.51  & 99.0    & -     & -     & -     & - \\
    RoBERTa \cite{roberta} & × & 2019  & -     & -     & 0.1   & 0     & 0.1   & 0.3 \\
    RotatE \cite{rotate} & × & 2019  & -     & -     & 29.0    & 23.4  & 32.2  & 39.0 \\
    QuatE \cite{quate} & × & 2019  & -     & -     & 27.6  & 22.7  & 30.1  & 35.9 \\
    ComplEx-N3-RP \cite{complex-n3-rp} & × & 2021  & -     & \underline{99.8}  & -     & -     & -     & - \\
    \midrule
    KG-BERT \cite{kg-bert} & \checkmark & 2019  & \underline{1.47}  & 99.0    & -     & -     & -     & - \\
    StAR \cite{star}  & \checkmark & 2021  & 1.49  & 99.1  & -     & -     & -     & - \\
    LaSS \cite{lass}  & \checkmark & 2021  & 1.56  & 98.9  & -     & -     & -     & - \\
    KEPLER \cite{kepler} & \checkmark & 2021  & -     & -     & 21.0    & 17.3  & 22.4  & 27.7 \\
    SimKGC \cite{simkgc} & \checkmark & 2021  & -     & -     & 35.8  & 31.3  & 37.6  & \underline{44.1} \\
    LP-BERT \cite{lp-bert} & \checkmark & 2022  & \textbf{1.18}  & \textbf{100}   & -     & -     & -     & - \\
    KGT5 \cite{kgt5}  & \checkmark & 2022  & -     & -     & 33.6  & 28.6  & 36.2  & 42.6 \\
    CSPromp-KG \cite{csprom-kg} & \checkmark & 2023  & -     & -     & \underline{38.0}    & \underline{34.3}  & \underline{39.9}  & \textbf{44.6} \\
    ReSKGC \cite{reskgc} & \checkmark & 2023  & -     & -     & \textbf{39.6}  & \textbf{37.3}  & \textbf{41.3}  & 43.7 \\
    \bottomrule
    \end{tabular}%
  }
  \caption{Experimental statistics of UMLS and Wikidata5M datasets under the LP subtask in transductive setting.}
  \label{tab:lp_umls_wikidata5m}%
\end{table}%

\begin{table}[!htbp]
  \centering
  \scalebox{0.88}{
    \begin{tabular}{c|cc|cccc}
    \toprule
    \multirow{2}[1]{*}{Model} & \multirow{2}[1]{*}{LLM-Enhanced} & \multirow{2}[1]{*}{Year} & \multicolumn{4}{c}{FB15k-237N} \\
          &       &       & MRR   & Hits@1 & Hits@3 & Hits@10 \\
    \midrule
    TransE \cite{fb15k} & × & 2013  & 25.5  & 15.2  & 30.1  & 45.9 \\
    DistMult \cite{distmult} & × & 2014  & 20.9  & 14.3  & 23.4  & 33.0 \\
    ComplEx \cite{complex} & × & 2016  & 24.9  & 18.0    & 27.6  & 38.0 \\
    ConvE \cite{wn18rr_umls_yago3-10} & × & 2018  & 27.3  & 19.2  & 30.5  & 42.9 \\
    RotatE \cite{rotate} & × & 2019  & 27.9  & 17.7  & 32.0    & 48.1 \\
    CompGCN \cite{compgcn} & × & 2020  & 31.6  & 23.1  & 34.9  & 48.0 \\
    \midrule
    KG-BERT \cite{kg-bert} & \checkmark & 2019  & 20.3  & 13.9  & 20.1  & 40.3 \\
    MTL-KGC \cite{mtl-kgc} & \checkmark & 2020  & 24.1  & 16.0    & 28.4  & 43.0 \\
    GenKGC \cite{genkgc} & \checkmark & 2022  & -     & 18.7  & 27.3  & 33.7 \\
    KG-S2S \cite{kgs2s} & \checkmark & 2022  & 35.4  & \underline{28.5}  & 38.8  & 49.3 \\
    CSPromp-KG \cite{csprom-kg} & \checkmark & 2023  & \underline{36.0}    & 28.1  & \underline{39.5}  & \underline{51.1} \\
    CD \cite{cd}    & \checkmark & 2024  & \textbf{37.2}  & \textbf{28.8}  & \textbf{41.0}    & \textbf{53.0} \\
    \bottomrule
    \end{tabular}%
  }
  \caption{Experimental statistics of FB15k-237N dataset under the LP subtask in transductive setting.}
  \label{tab:lp_fb15k-237n}%
\end{table}%

\begin{table}[!htbp]
  \centering
  \scalebox{0.88}{
    \begin{tabular}{c|cc|ccccc}
    \toprule
    \multirow{2}[1]{*}{Model} & \multirow{2}[1]{*}{LLM-Enhanced} & \multirow{2}[1]{*}{Year} & \multicolumn{5}{c}{NELL-One} \\
          &       &       & N-Shot & MRR   & Hits@1 & Hits@5 & Hits@10 \\
    \midrule
    GMatching \cite{nell-one} & × & 2018  & Five-Shot & 20    & 14    & 26    & 31 \\
    GMatching \cite{nell-one} & × & 2018  & One-Shot & 19    & 12    & 26    & 31 \\
    MetaR \cite{metar} & × & 2019  & Five-Shot & 26    & 17    & 35    & 44 \\
    MetaR \cite{metar} & × & 2019  & One-Shot & 25    & 17    & 34    & 40 \\
    StAR \cite{star}  & \checkmark & 2021  & Zero-Shot & 26    & 17    & 35    & 45 \\
    KG-S2S \cite{kgs2s} & \checkmark & 2022  & Zero-Shot & 31    & 22    & 41    & 48 \\
    \bottomrule
    \end{tabular}%
  }
  \caption{Experimental statistics under the LP subtask in inductive setting on NELL-One dataset.}
  \label{tab:lp_nellone}
\end{table}%

\begin{table}[!htbp]
  \centering
  \scalebox{0.88}{
    \begin{tabular}{c|cc|cccccc}
    \toprule
    \multirow{2}[1]{*}{Model} & \multirow{2}[1]{*}{LLM-Enhanced} & \multirow{2}[1]{*}{Year} & \multicolumn{6}{c}{Wikidata5M} \\
          &       &       & N-Shot & MRR   & MR    & Hits@1 & Hits@3 & Hits@10 \\
    \midrule
    DKRL \cite{dkrl}  & × & 2016  & Zero-Shot & 23.1  & \underline{78}    & 5.9   & 32.0    & 54.6 \\
    RoBERTa \cite{roberta} & × & 2019  & Zero-Shot & 7.4   & 723   & 0.7   & 1.0     & 19.6 \\
    KEPLER \cite{kepler} & \checkmark & 2021  & Zero-Shot & \underline{40.2}  & \textbf{28}    & \underline{22.2}  & \underline{51.4}  & \underline{73.0} \\
    SimKGC \cite{simkgc} & \checkmark & 2021  & Zero-Shot & \textbf{71.4}  & -     & \textbf{50.9}  & \textbf{78.5}  & \textbf{91.7} \\
    \bottomrule
    \end{tabular}%
  }
  \caption{Experimental statistics under the LP subtask in inductive setting on Wikidata5M dataset.}
  \label{tab:lp_wikidata5m}
\end{table}%

As shown in Table \ref{tab:lp_wn18rr} and \ref{tab:lp_fb15k-237}, early translation-based and semantic-matching methods can achieve respectable performance on benchmarks such as WN18RR and FB15k-237. For example, DistMult reports an MRR of 44.4\% and Hits@10 of 50.4\% on WN18RR, while TransE reaches around 47.4\% in Hits@10 on FB15k-237. However, these structurally guided models struggle with complex relational patterns, often reflected in lower Hits@1 (commonly only 4--20\% on WN18RR). Neural extensions like ConvE and R-GCN improve feature expressiveness; for instance, ConvE achieves an MRR of 31.2\% on FB15k-237, outperforming simpler baselines (e.g., TransE with MRR 27.9\%). Yet, they still face challenges on sparse graphs or large-scale data.

Recent approaches incorporating entity descriptions and prompts via large pre-trained encoders show further gains. SimKGC achieves an MRR of 66.7\% (Hits@10 = 80.5\%) on WN18RR, and CP-KGC reports MRR = 67.3\%, Hits@10 = 80.4\%. Both substantially surpass purely structural ComplEx, which has MRR around 44.9\%. On FB15k-237, methods such as OpenWorld KGC and Pretrain-KGE reach MRR values of roughly 30--35\%, indicating noticeable improvement over older baselines.

Encoder-decoder and decoder-based generative models exhibit mixed success. KGT5 attains an MRR of 34.3\% on FB15k-237. Although this outperforms many classical methods, it can trail behind strong retrieval-style methods like SimKGC. On the other hand, prompt-based generative approaches (e.g., CSPromp-KG) leverage textual cues effectively, pushing MRR to around 35--36\% on FB15k-237 and 38\% on Wikidata5M.

As listed in Table \ref{tab:lp_umls_wikidata5m}, UMLS is a domain-focused, smaller graph where structural baselines already score high Hits@10 (98--99\%). LP-BERT, by leveraging textual descriptions, further reduces mean rank to 1.18 and achieves Hits@10 of 100\%. In contrast, Wikidata5M is considerably more diverse: DistMult here only reaches an MRR of 25.3\%, while LLM-enhanced SimKGC and CSPromp-KG push the MRR to 35.8\% and 38.0\%, respectively. These improvements underscore how large language models can enrich entity representations, particularly under high relational heterogeneity.

As shown in Table \ref{tab:lp_fb15k-237n}, \ref{tab:lp_nellone}, and \ref{tab:lp_wikidata5m}, a more stringent setting involves inductive or low-data evaluation. For NELL-One, purely structural methods typically yield MRRs below 25\% in zero-/few-shot modes. By contrast, text-enhanced KG-S2S attains 31\% MRR under zero-shot. Likewise, SimKGC on the zero-shot split of Wikidata5M achieves an MRR of 71.4\%, outperforming baseline encoders (e.g., RoBERTa with only 7.4\%). Such findings illustrate the strong inductive capacity of LLM-based techniques to handle unseen entities/relations, thanks to rich textual pre-training.

Overall, LLM-enhanced methods often surpass or closely match purely structural approaches, especially on large or more complex graphs. In zero-shot scenarios, their reliance on prior linguistic knowledge further drives performance gains. Nevertheless, a uniform advantage is not guaranteed. Certain encoder-decoder methods, although adept at text generation, can be less effective at large-scale structural inference compared to advanced graph-based retrieval strategies. Moreover, some of the latest decoder-based large language models lack comprehensive experimental results on all benchmarks, hindering fair one-to-one comparisons. Future work may expand empirical coverage, refine prompt engineering, and strike a balance between accuracy and feasible inference costs.

\section{Future Directions} \label{Sec_5}
Recent advances in LLMs mark a significant inflection point in KRL research, presenting exciting opportunities for future exploration of effectively integrating the advantages of LLMs into KRL. We propose the following six recommendations:

\subsection{Dynamic and Multimodal Knowledge Representation}
As knowledge evolves far beyond static triples, future KRL approaches must accommodate temporal or event-based representations that can handle the dynamic and evolving characteristics inherent in real-world data. Modeling sequences of events and capturing causal dependencies enables a more fine-grained understanding of how entities and relations form, merge, or dissolve over time. Such temporal awareness can reveal nuanced patterns that remain invisible under static assumptions, thereby enhancing both accuracy and timeliness of knowledge graphs.

Adopting a temporal perspective can also highlight previously overlooked dynamics in evolving entities and their interactions. For instance, an entity might gain or lose attributes, or a relationship might change direction based on new observations. These temporal updates not only demand continuous monitoring of knowledge states but also require mechanisms for effectively reconciling new information with existing graph structures.

Beyond temporal factors, extending KRL to multimodal data such as text, images, and videos further enriches the scope of knowledge graphs. However, fusing different data types in a cohesive representation presents technical challenges, including the risk of losing details specific to each modality. Overcoming these obstacles often calls for specialized modeling architectures and benchmarking strategies that explicitly measure performance in dynamic, multimodal contexts.

\subsection{Efficiency, Robustness, and Explainability}
Although LLMs provide significant improvements for representing knowledge, the accompanying computational overhead restricts their straightforward deployment in numerous real-world scenarios. Reducing model size, optimizing inference time, and carefully allocating resources become critical steps for practical adoption. Techniques such as model pruning and quantization, along with knowledge distillation into more compact architectures, help strike a balance between performance and efficiency.

Robustness against noise and domain shifts is equally important in real-world knowledge contexts. Domain adaptation, adversarial training, and probabilistic representations can help LLM-enhanced KRL maintain stable performance even when new or contradictory information is introduced. Greater interpretability complements these efforts by increasing user trust. Transparent reasoning pipelines, attention visualization, and causal interpretability methods all contribute to a clearer understanding of how and why the system reaches a particular conclusion.

\subsection{Downstream Tasks with KRL+LLM}
Integrating KRL with LLM can significantly enhance the ability to solve various problems and downstream tasks. Future research should explore how these models can be utilized to improve human decision-making and knowledge discovery in different domains.

Interactive systems enable humans to query and interact with LLM-enhanced KRL models that enhance knowledge exploration and problem-solving. The natural language interface allows users to ask questions and receive explanations, thus making complex knowledge graphs more straightforward to understand.

Research could also develop collaborative learning frameworks that incorporate continuous human feedback into the learning process of the model. Techniques such as active learning can improve understanding of the model. Promoting a symbiotic relationship between humans and AI can capitalize on the strengths of both parties to achieve superior performance.

\subsection{Pre-LLM Models and Data Augmentation}
The integration of LLMs into KRL goes far beyond merely improving the understanding of textual information. Future research should explore a broader range of collaborative enhancements that can be realized by leveraging the strengths of LLMs in conjunction with traditional approaches. One promising way is the development of mixed frameworks that allow pre-LLM models, which are often more lightweight and efficient, to work collaboratively with LLMs. These models could handle more straightforward or routine tasks, while LLMs focus on more complex reasoning and representation tasks. This division of labour can significantly enhance both the performance and efficiency of KRL, ensuring that the strengths of various types of models are used in the most appropriate contexts. 

Moreover, data augmentation offers another layer of improvement by expanding the diversity and richness of training data. This can involve generating synthetic data from existing datasets, creating variations of textual, visual, or multimodal inputs. Such augmented data can help pre-LLM models and LLM-enhanced models alike, enabling better generalization and robustness across different domains.

\subsection{Graph-Centric Instruction Tuning}
Another promising angle is to adapt advanced LLMs to graph-centric contexts via specialized prompts and instruction-tuning procedures. By designing domain-relevant templates that encapsulate the structure and semantics of knowledge graphs, LLMs can be primed to deliver consistent and precise outputs aligned with triple-oriented representations. This tailored prompting strategy may also help constrain the inherently open-ended nature of LLM responses, which is especially important for tasks that require well-defined, structured outputs.

Instruction fine-tuning can further bridge the gap between textual comprehension and formal graph outputs. Extending sequence-to-sequence or decoder-based LLMs to generate or update triples directly, rather than merely producing a textual summary, allows for a smoother integration of large-scale pre-trained knowledge with KRL tasks. This strategy holds promise for applications ranging from data validation to large-scale knowledge graph construction.

\subsection{Incremental and Continual Learning}
Finally, incremental and continual learning mechanisms address the reality that knowledge is never static. To remain current without retraining from scratch, LLM-enhanced KRL models must absorb new facts and relationships as they emerge, minimizing downtime and computational overhead. Employing an on-the-fly update architecture, combined with streaming data paradigms, allows knowledge graphs to reflect recent changes while still retaining core learned patterns.

Preserving previously acquired knowledge becomes crucial in settings where frequent updates can otherwise destabilize the learned model. Regularization-based strategies and replay mechanisms help mitigate catastrophic forgetting by ensuring that newer information does not overwrite earlier insights. Such solutions become especially significant in domains where continuity and institutional memory are key to sustaining accurate, context-rich representations of the underlying data.

\section{Conclusion} \label{Sec_6}
Enhancing Knowledge Representation Learning (KRL) with Large Language Models (LLMs) is an emerging research area that has garnered significant attention in both academia and industry. This article presents a comprehensive overview of recent advancements in this field. It begins by introducing the foundational work on KRL prior to the enhancement by LLMs, highlighting traditional methods and their limitations. Next, existing LLM-enhanced KRL methods are categorized, focusing on encoder-based, encoder-decoder-based, and decoder-based techniques, and a taxonomy is provided based on the variety of downstream tasks they address. Finally, the paper discusses future research directions in this rapidly evolving domain.

\backmatter

\bmhead{Acknowledgements}
This work is supported by the National Natural Science Foundation of China (62472311, U23B2057, 62176185).

\bibliography{sn-bibliography}

\end{document}